%% file: main.tex
\documentclass{article}

\usepackage{microtype}
\usepackage{enumitem}
\usepackage{graphicx,svg,tikz}
\usepackage{subcaption}
\usepackage{booktabs} % for professional tables

\usepackage{hyperref}
\usepackage{xurl}

\usepackage[accepted]{icml2025}

\usepackage{mathtools}
\usepackage{amsthm}

\usepackage[capitalize,noabbrev]{cleveref}

\theoremstyle{plain}

\theoremstyle{definition}

\theoremstyle{remark}

\input{math_cmd.tex}

\icmltitlerunning{Training Dynamics of In-Context Learning in Linear Attention}

\begin{document}

\twocolumn[
\icmltitle{Training Dynamics of In-Context Learning in Linear Attention}

\icmlsetsymbol{equal}{*}

\begin{icmlauthorlist}
\icmlauthor{Yedi Zhang}{gatsby}
\icmlauthor{Aaditya K. Singh}{gatsby}
\icmlauthor{Peter E. Latham}{gatsby,equal}
\icmlauthor{Andrew Saxe}{gatsby,swc,equal}
\end{icmlauthorlist}

\icmlaffiliation{gatsby}{Gatsby Computational Neuroscience Unit, University College London}
\icmlaffiliation{swc}{Sainsbury Wellcome Centre, University College London}

\icmlcorrespondingauthor{Yedi Zhang}{yedi@gatsby.ucl.ac.uk}

% You may provide any keywords that you find helpful for describing your paper; these are used to populate the "keywords" metadata in the PDF but will not be shown in the document
\icmlkeywords{In-Context Learning, Linear Attention, Learning Dynamics, Gradient Flow}

\vskip 0.3in
]



\printAffiliationsAndNotice{\icmlEqualContribution} % otherwise use the standard text.

\begin{abstract}
While attention-based models have demonstrated the remarkable ability of in-context learning (ICL), the theoretical understanding of how these models acquired this ability through gradient descent training is still preliminary. Towards answering this question, we study the gradient descent dynamics of multi-head linear self-attention trained for in-context linear regression. We examine two parametrizations of linear self-attention: one with the key and query weights merged as a single matrix (common in theoretical studies), and one with separate key and query matrices (closer to practical settings). For the merged parametrization, we show that the training dynamics has two fixed points and the loss trajectory exhibits a single, abrupt drop. We derive an analytical time-course solution for a certain class of datasets and initialization. For the separate parametrization, we show that the training dynamics has exponentially many fixed points and the loss exhibits saddle-to-saddle dynamics, which we reduce to scalar ordinary differential equations. During training, the model implements principal component regression in context with the number of principal components increasing over training time. Overall, we provide a theoretical description of how ICL abilities evolve during gradient descent training of linear attention, revealing abrupt acquisition or progressive improvements depending on how the key and query are parametrized.
\end{abstract}

\section{Introduction}
Self-attention-based models, such as transformers \citep{vaswani17transformer}, exhibit a remarkable ability known as in-context learning \citep{brown20fewshot}. That is, these models can solve unseen tasks based on exemplars in the context of an input prompt. In-context learning (ICL) is critical to the flexibility of large language models, allowing them to solve tasks not explicitly included in their training data. However, it remains unclear how architectures like self-attention acquire this ability through gradient descent training.

Seminal work by \citet{anthropic22ih} identified an intriguing trait in the training dynamics of ICL: the ICL ability often emerges abruptly, coinciding with an abrupt drop in loss during training. This abrupt learning phase can reflect the formation of an induction head in the ICL setting \citep{anthropic22ih,reddy24abrupt,aaditya24induction,edelman24markov}, and can also occur more broadly in transformer training dynamics \citep{nanda23grok,chen24sudden,hoffmann24eureka}.
Furthermore, \citet{aaditya23transient} found that ICL may often be a transient ability that the transformers acquire and then lose over the course of long training time, a phenomenon that has since been reproduced in many settings \citep{he24grok,anand24dual,chan24icliwl,nguyen24kinetics,park24competition,aaditya25coopetition}.
These findings underscore the importance of understanding not only the ICL ability in trained models, but its full training dynamics.

This work aims to provide a theoretical description of how the ICL ability evolves in gradient descent training.
To do so, we consider the increasingly common setup of linear attention\footnote{We refer to linear self-attention as linear attention in this paper.} \citep{oswald23GD} trained on an in-context linear regression task \citep{garg22function}.
The in-context linear regression task, in which the model needs to perform linear regression on the data in context, is a canonical instantiation of ICL \citep{garg22function,akyurek23algorithm,oswald23GD,ahn23preconditioned,bai23statistician}. 
The linear attention model, which has been used in many prior studies \citep{schlag21lintransformer,oswald23GD,ahn23preconditioned,zhang24jmlr,wu24howmany,fu24convergence,mahankali24linear,duraisamy24finite,li24finegrain,yau24polytime,cengiz24asymptotic,frei24benign}, reproduces key optimization properties of practical transformers \citep{ahn24linear} and is more amenable to theoretical analysis. 
Importantly, despite its name, linear attention is a nonlinear model, as it removes the softmax operation but is still a nonlinear function of the input.

We study two common parametrizations of multi-head linear attention: (i) $\attn_{\text M}$, linear attention where the key and query matrices in each head are merged into a single matrix, a reparametrization procedure widely used in theoretical studies on transformers \citep{ahn23preconditioned,tian23snap,ataee23maxmargin,zhang24jmlr,zhang24mlp,siyu24dynamics,wu24howmany,kim24meanfield,huang24convergence,wang24sparse,ildiz24markov,ren24bigram,tarzanagh24svm,vasudeva24implicit,cengiz24asymptotic,chen24learnability,julistiono24mirror,yau24polytime,anwar24adversarial,huang25CoT}; (ii) $\attn_{\text S}$, linear attention with separate key and query matrices, which is closer to the implementation of attention in real-world transformers \citep{vaswani17transformer}.
We specify the fixed points in the loss landscapes, as well as how gradient descent training dynamics traverses the landscape. Our findings are summarized as follows.
\vspace{-1.5ex}
\begin{itemize}[leftmargin=*,itemsep=0.5pt]
    \item We find two fixed points in the training dynamics of $\attn_{\text M}$, and exponentially many fixed points in that of $\attn_{\text S}$.
    \item We show a single, abrupt loss drop in training $\attn_{\text M}$ from small initialization and derive an analytical time-course solution when the input token covariance is white. We show saddle-to-saddle training dynamics in training $\attn_{\text S}$ from small initialization and reduce the high-dimensional training dynamics to scalar ordinary differential equations through an ansatz. We demonstrate the rank of the separate key and query weights affects the dynamics by shortening the duration of certain plateaus.
    \item We identify the in-context algorithm of the converged and early stopped models. When $\attn_{\text M}$ and $\attn_{\text S}$ are trained to convergence, they approximately implement least squares linear regression in context. When the training of $\attn_{\text S}$ early stops during the $(m+1)$-th loss plateau, it approximately implements principal component regression in context with the first $m$ principal components.
    \item As a tool for our analysis, we show that when trained on in-context linear regression tasks, $\attn_{\text M}$ is equivalent to a two-layer fully-connected linear network with a cubic feature map as input, and $\attn_{\text S}$ is equivalent to a sum of three-layer convolutional linear networks with the same cubic feature map as input.
    \item We empirically demonstrate that the single and multiple loss drops also occur in softmax $\attn_{\text M}$ and $\attn_{\text S}$, respectively.
\end{itemize}
\vspace{-1.5ex}
Comparing the two models, we find that the ICL ability evolves differently in them: $\attn_{\text M}$ acquires the in-context linear regression ability through one abrupt loss drop, while $\attn_{\text S}$ acquires this ability by \textit{progressively improving} on in-context principal component regression. This makes a theoretical case for the progressive improvements of ICL in gradient descent training. Our results also reveal how parametrization, such as merged versus separate key and query and the rank of the separate key and query weights, influences the loss landscape and training dynamics. This motivates future research to take the parametrization factor into account when studying the landscape and dynamics of attention models.

\vspace{-1ex}
\section{Preliminaries}
\textbf{Notation.}
Non-bold small and capital symbols are scalars. Bold small symbols are column vectors. Bold capital symbols are matrices. $\|\cdot\|$ denotes the $\ell^2$ norm of a vector or the Frobenius norm of a matrix. $\VEC(\cdot)$ represents flattening a matrix to a column vector by stacking its columns. 
For example, $\VEC\begin{bmatrix}1&3\\2&4\end{bmatrix} 
= \begin{bmatrix}1&2&3&4\end{bmatrix}^\T$.
% We use $i=1,\cdots,H$ to denote the index of an attention head, $\mu=1,\cdots,P$ to denote
We use $i=1,\cdots,H$ to denote the index of an attention head, $\mu=1,\cdots,P$ to denote the index of a training sample, and $n=1,\cdots,N$ to denote the index of a token in a sample.

\vspace{-1ex}
\subsection{In-Context Linear Regression Task  \label{sec:def-data}}
We study a standard ICL task of predicting the next token. The input is a sequence $\{ \vx_1, y_1, \vx_2, y_2, \cdots, \vx_N, y_N, \vx_q \}$ and the desired output is $y_q$. We refer to $\vx_q$ as the query token, $\{ \vx_1, y_1, \vx_2, y_2, \cdots, \vx_N, y_N\}$ as the context, and $N$ as the context length. By convention \citep{ahn23preconditioned,zhang24jmlr,zhang24mlp,siyu24dynamics,huang24convergence}, the input sequence is presented to the model as a matrix $\mX$, defined as
\begin{align}
\mX = \begin{bmatrix}
\vx_1 & \vx_2  & \cdots & \vx_N & \vx_q \\
y_1 & y_2 & \cdots & y_N & 0
\end{bmatrix} \in \sR^{(D+1)\times(N+1)} ,
\end{align}
where $\vx_1,\cdots,\vx_N,\vx_q \in \sR^D$ and $y_1,\cdots,y_N \in \sR$.

We are given a training dataset $\left\{ \mX_\mu, y_{\mu,q} \right\}_{\mu=1}^P$ consisting of $P$ samples.
All $\vx$ tokens are independently sampled from a $D$-dimensional zero-mean normal distribution with covariance $\mLambda$,
\begin{align}
\vx_{\mu,n}, \vx_{\mu,q} \sim \mathcal N(\vzero, \mLambda), n = 1,\cdots,N , \mu = 1,\cdots,P .
\hspace{-0.1ex}
\end{align}
We consider the in-context linear regression task, where the $y_n$ in context and the target output $y_q$ are generated as a linear map of the corresponding $\vx_n$ and $\vx_q$ \citep{garg22function}. For each sequence $\mX_\mu$, we independently sample a task vector $\vw_\mu$ from a $D$-dimensional standard normal distribution, $\vw_\mu \sim \mathcal N(\vzero, \mI)$, and generate $y_{\mu,n} = \vw_\mu^\T \vx_{\mu,n} ,
y_{\mu,q} = \vw_\mu^\T \vx_{\mu,q} , n = 1,\cdots,N ,\, \mu = 1,\cdots,P$.
Note that the task vector $\vw_\mu$ is fixed for all tokens in one sample sequence but varies across different samples, and is independent of the tokens $\vx_{\mu,1}, \cdots, \vx_{\mu, N}, \vx_{\mu,q}$.

\vspace{-1ex}
\subsection{Multi-Head Self-Attention}
\vspace{-0.5ex}
A standard multi-head softmax self-attention layer \citep{vaswani17transformer} takes the matrix $\mX$ as input and returns a matrix of the same size,
\begin{align*}
\attn (\mX) = \mX + \sum_{i=1}^H \mW_i^V \mX \mathsf{smax} \left( \frac{\mX^\T {\mW_i^K}^\T \mW_i^Q \mX}{\rho} \right)
\end{align*}
where $H$ is the number of heads, $\rho$ is a scaling factor, and $\mW_i^V, \mW_i^K, \mW_i^Q$ are the trainable value, key, and query matrices in the $i$-th head.
The prediction for $y_q$ is the bottom right entry of the output matrix:
\begin{align}
\hat y_q = \attn (\mX)_{D+1,N+1} .
\end{align}
In this work, we consider multi-head linear self-attention, where we remove the softmax operation and take $\rho=N$.
Specifically, we study two common parametrizations of linear attention: (i) linear attention with merged key and query introduced in \cref{sec:attnM-def} and analyzed in \cref{sec:attnM}; (ii) linear attention with separate key and query introduced in \cref{sec:attnS-def} and analyzed in \cref{sec:attnS,sec:attnS-lowrank}.

\subsection{Linear Attention with Merged Key and Query  \label{sec:attnM-def}}
The multi-head linear attention $\attn_{\text M}$ with the key and query matrices in each head merged as a single matrix ${\mW^K_i}^\T \mW^Q_i = \mW^{KQ}_i$ computes
\begin{align*}
\attn_{\text M}(\mX) &= \mX + \sum_{i=1}^H \frac1N \mW^V_i \mX \mX^\T \mW^{KQ}_i \mX ,
\end{align*}
where the terms can be written in block form,
\begin{align*}
\mX \mX^\T &= \begin{bmatrix}
\left(\vx_q \vx_q^\T + \sum_{n=1}^N \vx_n \vx_n^\T \right) & \sum_{n=1}^N \vx_n y_n  \\
\sum_{n=1}^N y_n \vx_n^\T & \sum_{n=1}^N y_n^2 \end{bmatrix} ,
\end{align*}
and 
\begin{align*}
\mW^V_i &= \begin{bmatrix}
* & * \\ \vv_i^\T & v_i
\end{bmatrix} ,
\mW^{KQ}_i = \begin{bmatrix}
\mU_i & * \\ \vu_i^\T & *
\end{bmatrix} .
\end{align*}
The blocks have dimensionalities $\vv_i,\vu_i \in \sR^D, v_i \in \sR , \mU_i \in \sR^{D\times D}$. The $*$ blocks denote entries that do not contribute to the computation of $\attn (\mX)_{D+1,N+1}$. 
% With the block matrix notations, the bottom right entry of $\attn_{\text M}(\mX)$ is
% \begin{align}    \label{eq:rawdef-attnM}
% \attn_{\text M} (\mX)_{D+1,N+1}
% = \sum_{i=1}^H \begin{bmatrix}
% \vv_i^\T & v_i
% \end{bmatrix}
% \frac{\mX \mX^\T}{N}
% \begin{bmatrix}
% \mU_i \\ \vu_i^\T
% \end{bmatrix} \begin{bmatrix}
% \vx_q \\ 0
% \end{bmatrix} ,
% \end{align}
Following \citet{ahn23preconditioned,zhang24jmlr,kim24meanfield,huang24convergence}, we initialize $\vv_i,\vu_i=\vzero$ as they are not required for this model to achieve global minimum loss on the in-context linear regression task. When $\vv_i$ and $\vu_i$ are initialized to zero, they will remain zero throughout training (see \cref{supp:attnM-zerouv}). With the reduction $\vv_i,\vu_i=\vzero$, the prediction for $y_q$, which is the bottom right entry of $\attn_{\text M}(\mX)$, is
\begin{align}  \label{eq:def-attnM}
\attn_{\text M} (\mX)_{D+1,N+1}
= \sum_{i=1}^H v_i \vbeta^\T \mU_i \vx_q  ,  \tag{M}
\end{align}
where $\vbeta$ is the correlation between $\vx_n$ and $y_n$ in context,
\begin{align}  \label{eq:def-beta}
\vbeta \equiv \frac1N \sum_{n=1}^N y_n \vx_n  .
\end{align}

\subsection{Linear Attention with Separate Key and Query  \label{sec:attnS-def}}
% In multi-head attention with separate key and query matrices, we follow the standard practice \citep{vaswani17transformer} of using low-rank key and query matrices where the rank $R\leq D$. Additionally, we enforce $RH\geq D$ to prevent expressivity limitations from affecting the behaviors we study.\footnote{In practice, usually $RH=D$.}
In multi-head attention with separate key and query, we follow the standard practice \citep{vaswani17transformer} of using low-rank key and query matrices where the rank $R\leq D$ and $RH\geq D$. In practice, usually $RH=D$.
The multi-head linear attention $\attn_{\text S}$ with separate rank-$R$ key and query matrices computes
\begin{align*}
\attn_{\text S}(\mX) &= \mX + \sum_{i=1}^H \frac1N \mW^V_i \mX \mX^\T {\mW_i^K}^\T \mW_i^Q \mX .
\end{align*}
We can write the value, key, and query weights in block form,
\begin{align*}
\mW^V_i = \begin{bmatrix}
* & * \\ \vv_i^\T & v_i
\end{bmatrix} , 
\mW_i^K = \begin{bmatrix}
\vk_{i,1}^\T & k_{i,1}  \\ 
\vdots & \vdots  \\ 
\vk_{i,R}^\T & k_{i,R}
\end{bmatrix} , 
\mW_i^Q = \begin{bmatrix}
\vq_{i,1}^\T & *  \\ 
\vdots & \vdots  \\ 
\vq_{i,R}^\T & *
\end{bmatrix}  .
\end{align*}
The blocks have dimensionalities $v_i,k_{i,r} \in \sR$ and $\vv_i,\vk_{i,r},\vq_{i,r} \in \sR^D$ $(r=1,\cdots,R)$.
Similarly to the case with merged key and query, we initialize $\vv_i=\vzero,k_{i,r}=0$; they will remain zero throughout training (see \cref{supp:attnS-lowrank-zerouv}). 
With $\vv_i=\vzero$ and $k_{i,r}=0$, the multi-head linear attention with separate rank-one key and query matrices computes
\begin{align}  \label{eq:def-attnS}
\attn_{\text S} (\mX)_{D+1,N+1} = \sum_{i=1}^H \sum_{r=1}^R v_i \vbeta^\T \vk_{i,r} \vq_{i,r}^\T \vx_q  ,  \tag{S}
\end{align}
where $\vbeta$ is the input-output correlation in context defined in \cref{eq:def-beta}. The expression of \cref{eq:def-attnS} already reveals interesting insight. It implies that linear attention with $H$ heads and rank-$R$ key and query differs from linear attention with $RH$ heads and rank-one key and query only in the sharing of certain value weights.

\subsection{Gradient Flow Training Dynamics}
We train the linear attention model using gradient descent on squared loss of the query token\footnote{We can also handle next token prediction loss (\cref{supp:varylen}).}, that is $\Ls = \E (y_q - \hat y_q)^2$.
% \begin{align} \label{eq:loss}
% \Ls = \lim_{P\to \infty} \frac1P \sum_{\mu=1}^P (y_{\mu,q} - \hat y_{\mu,q})^2 = \E (y_q - \hat y_q)^2  \, .
% \end{align}
We analyze the gradient flow dynamics on the loss, given by
\begin{align}  \label{eq:grad-flow}
\tau \frac{\diff \mW}{\diff t} = - \frac12 \frac{\partial \Ls}{\partial \mW} = \E \left[ (y_q - \hat y_q) \frac{\partial \hat y_q}{\partial \mW} \right]  ,
\end{align}
where $\tau$ is the time constant. The gradient flow dynamics captures the behavior of gradient descent in the limit of a small learning rate.

\begin{figure*}
  \centering
  \includegraphics[height=3.3cm]{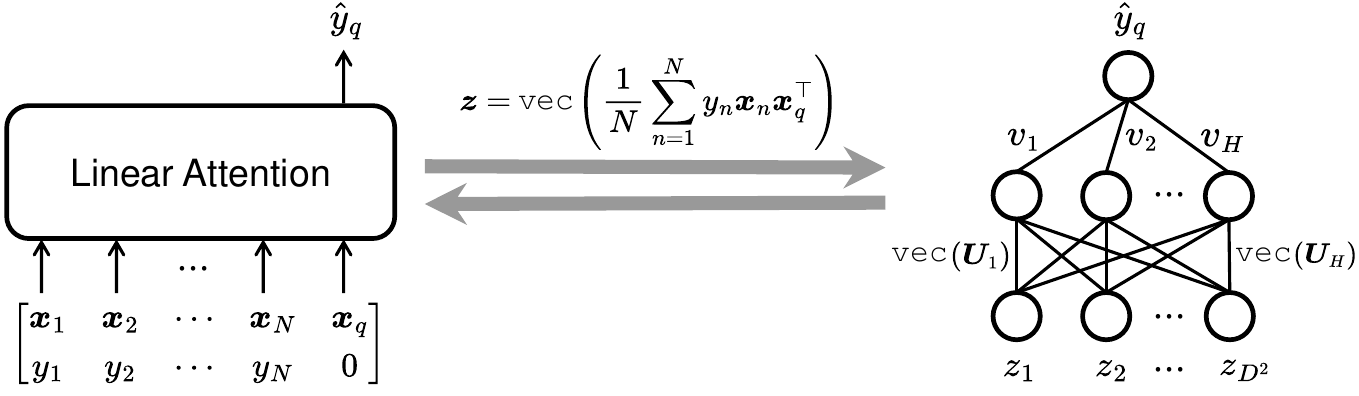}
  \hfill
  \includegraphics[height=3.3cm]{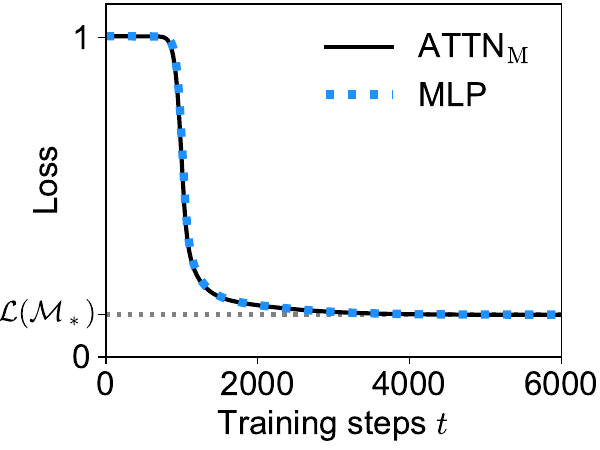}
  \vspace{-1ex}
  \caption{Multi-head linear attention with merged key and query, $\attn_{\text M}(\mX)_{D+1,N+1}$, is equivalent to a two-layer fully-connected linear network with cubic feature input, $\mathsf{MLP} (\vz)$. \textit{Left}: Schematic of the equivalence. 
  \textit{Right}: Loss trajectories of linear attention and the fully-connected linear network match well. The two models are trained with the same data and initialization. Both exhibit the characteristic abrupt loss drop documented by prior work on the ICL dynamics in linear \citep{oswald23GD} and softmax attention \citep{aaditya24induction}. Here $D=4,N=31,H=8$.}
  \label{fig:attnM}
\end{figure*}

\section{Linear Attention with Merged Key and Query \label{sec:attnM}}
We first study multi-head linear attention with the key and query matrices merged as a single matrix, as described by \cref{eq:def-attnM}.

\subsection{Connection to A Fully-Connected Linear Network  \label{sec:attnM-mlp}}
The $H$-head linear attention with input sequence $\mX$ defined in \cref{eq:def-attnM} can be viewed as a two-layer width-$H$ fully-connected linear network with a cubic feature $\vz(\mX)$ as input,
\begin{align}  \label{eq:attnM-mlp}
\attn_{\text M} (\mX)_{D+1,N+1} &= \sum_{i=1}^H v_i \vbeta^\T \mU_i \vx_q  \nonumber\\
&= \sum_{i=1}^H v_i \VEC(\mU_i)^\T \VEC \left( \vbeta \vx_q^\T \right)  \nonumber\\
&= \vw_2^\T \mW_1 \vz = \mathsf{MLP} (\vz)  ,
\end{align}
where 
\begin{align}  \label{eq:def-feat-z}
\vw_2 = \begin{bmatrix}
v_1 \\ v_2 \\ \vdots \\ v_H
\end{bmatrix}
,
\mW_1 = \begin{bmatrix}
\VEC(\mU_1)^\T \\ \VEC(\mU_2)^\T \\ \vdots \\ \VEC(\mU_H)^\T
\end{bmatrix} 
,
\vz (\mX) = \VEC \left( \vbeta \vx_q^\T  \right)  .
\end{align}
% The middle line of \cref{eq:attnM-mlp} can be understood through the definition of the quadratic form $\vbeta^\T \mU_i \vx_q$.
The feature $\vz \in \sR^{D^2}$, whose entries are cubic functions of the entries in the original sequence $\mX$, is the input to the equivalent two-layer fully-connected linear network.
The stacked value weights correspond to the second-layer weights $\vw_2 \in \sR^H$ of the fully-connected linear network. The stacked merged key-query weights correspond to the first-layer weights $\mW_1 \in \sR^{H\times D^2}$ of the fully-connected linear network.
A schematic of this equivalence is given in \cref{fig:attnM}.

% The somewhat surprising and useful fact here is that the linear attention in this regime is equivalent to a \textit{fully-connected} linear network (with cubic features of $\mX$ as input). It is evident that the linear attention is a cubic function of $\mX$ given its definition in \cref{eq:def-attnM}, while it is less evident that the cubic function is a fully-connected linear network of $\vz(\mX)$. This equivalence draws a connection between the well-established fully-connected linear network and the more recent attention model, enabling us to apply the theoretical machinery of the former to the latter.

\subsection{Loss Landscape: Two Fixed Points  \label{sec:attnM-landscape}}
The gradient flow training dynamics of the linear attention or the equivalent two-layer fully-connected linear network given in \cref{eq:attnM-mlp} is
\begin{subequations}  \label{eq:attnM-gd}
\begin{align}
\tau \dot \mW_1 &= \vw_2 \left( \E \left(y_q \vz^\T \right) - \vw_2^\T \mW_1 \E \left(\vz \vz^\T \right) \right) ,  \label{eq:attnM-gd-W1} \\
\tau \dot \vw_2 &= \mW_1 \left( \E \left(y_q \vz^\T \right) - \vw_2^\T \mW_1 \E \left(\vz \vz^\T \right) \right)^\T .  \label{eq:attnM-gd-w2}
\end{align}
\end{subequations}
There are two manifolds of fixed points in this dynamical system: one is the unstable fixed point at zero, denoted $\gM_0$, and the other is a manifold of stable fixed points at the global minimum, denoted $\gM_*$,
\begin{subequations}  \label{eq:attnM-def-M}
\begin{align}
\gM_0 &= \{ \vw_2 = \vzero,  \mW_1 = \vzero \}   \label{eq:attnM-M0}\\
\gM_* &= \left\{ \vw_2, \mW_1 \big| \vw_2^\T \mW_1 = \E \left( y_q \vz^\T \right) \E \left( \vz \vz^\T \right)^{-1} \right\} \hspace{-0.7ex} \label{eq:attnM-Mstar}
\end{align}
\end{subequations}

\subsection{Training Dynamics: An Abrupt Drop in the Loss  \label{sec:attnM-dynamics}}
We have shown the linear attention defined in \cref{eq:def-attnM} is equivalent to a fully-connected linear network with cubic feature input. Since this equivalence holds at the level of the computation of the model, the equivalence applies to the training dynamics with any initialization and optimizer.
Here we discuss the training dynamics from small initialization, commonly referred to as the rich learning regime \citep{woodworth20kernel}.

With small initialization, the network is initially near the unstable fixed point, $\gM_0$, at zero. As training progresses, the network escapes from the unstable fixed point, and subsequently converges to a stable fixed point on the global minimum manifold, $\gM_*$.
The time it takes to escape from the unstable fixed point is approximately $\frac{\tau}{\| \mLambda^2 \|} \ln \frac1{w_\init}$, where the initialization scale $w_\init$ is the initial $\ell^2$ norm of a layer (see \cref{supp:attnM-duration}). Because the time to escape from the unstable fixed point starting from small initialization is long, the loss exhibits an initial plateau followed by an abrupt drop, as validated by simulations in \cref{fig:attnM}.
In particular, when the input token covariance is white $\mLambda=\mI$ and the initialization is infinitesimally small, we exploit the equivalence between linear attention and linear networks to derive an analytical time-course solution (see \cref{supp:attnM-solution}) and obtain 
\begin{align}  \label{eq:attnM-solution}
\attn_{\text M} (\mX;t)_{D+1,N+1} = \sigma (t) \vbeta^\T \vx_q  , \nonumber\\
\text{where }\, \sigma (t) = \frac{e^{2\sqrt D\frac t\tau}}{\left( 1+\frac{1+D}N \right) \left(e^{2\sqrt D\frac t\tau}-1\right)+\frac{\sqrt D}{w_\init^2}}  .
\end{align}

Since $\sigma (t)$ is a rescaled and shifted sigmoid function, the weights and the loss trajectories have sigmoidal shapes, characterized by a plateau followed by a rapid drop.

\subsection{ICL Algorithm: Least Squares Regression}
When the linear attention model converges to the global minimum manifold $\gM_*$ at the end of training, the model implements
\begin{align}  \label{eq:attnM-converged}
\begin{split}
\attn_{\text M} (\mX)_{D+1,N+1}  = \E \left( y_q \vz^\T \right) \E \left( \vz \vz^\T \right)^{-1} \vz  \\
= \vbeta^\T \left( \mLambda + \frac{\mLambda + \tr(\mLambda) \mI}N \right)^{-1} \vx_q  , 
\end{split} 
\end{align}
where the first equality follows directly from \cref{eq:attnM-mlp,eq:attnM-Mstar} and the second equality is proved in \cref{supp:attnM-converged}.
\cref{eq:attnM-converged} reveals an intriguing duality: the linear regression solution in the cubic feature space of $\vz$ is the in-context linear regression solution in the original space of the $\vx_n,y_n$ token pairs in a sequence $\mX$.
The first line of \cref{eq:attnM-converged} is the linear regression solution of fitting $y_{\mu,q}$ with $\vz_\mu$ for all training sequences $\mu=1,\cdots,P$.
The second line of \cref{eq:attnM-converged} is approximately the in-context linear regression solution, which fits $y_{\mu,n}$ with $\vx_{\mu,n} (n=1,\cdots,N)$ for each sequence $\mX_\mu$. When the sequence length $N$ is large, the model recovers the inverse of the true covariance matrix,
\begin{align*}
\lim_{N\to \infty}
\vbeta^\T \left( \mLambda + \frac{\mLambda + \tr(\mLambda) \mI}N \right)^{-1} \vx_q
= \vbeta^\T \mLambda^{-1} \vx_q .
\end{align*}
Here $\vbeta$ is the $\vx_n,y_n$ correlation in a sequence $\mX$, and $\mLambda$ is the covariance of all $\vx_n$ tokens in all training sequences, which approximates the covariance of $\vx_n$ in each individual sequence.

\section{Linear Attention with Separate Rank-One Key and Query \label{sec:attnS}}
We now study multi-head linear attention with separate low-rank key and query matrices. Because the rank-one case captures most of the behaviors of the general rank-$R$ case, we focus on the rank-one case in this section and defer the rank-$R$ case to \cref{sec:attnS-lowrank}. When $R=1$, the model definition in \cref{eq:def-attnS} simplifies to 
\begin{align}  \label{eq:def-attnS-rank1}
\attn_{\text S} (\mX)_{D+1,N+1} = \sum_{i=1}^H v_i \vbeta^\T \vk_i \vq_i^\T \vx_q  .
\end{align}

\subsection{Connection to Convolutional Linear Networks}
\begin{figure}
  \centering
  \includegraphics[width=0.75\linewidth]{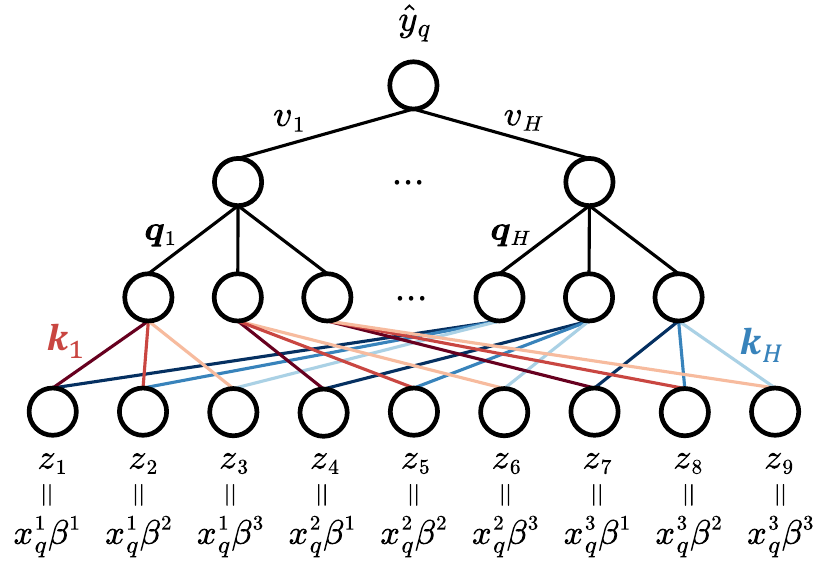}
  \vspace{-2ex}
  \caption{Multi-head linear attention with separate rank-one key and query $\attn_{\text S}(\mX)_{D+1,N+1}$ is a sum of $H$ (number of heads) three-layer convolutional linear networks with the cubic feature $\vz$ as input. Here we take $D=3$ to avoid clutter. Entries in the vectors are denoted as $\vx_q = \left[x_q^1, x_q^2, x_q^3 \right]^\T, \vbeta = \left[\beta^1, \beta^2, \beta^3 \right]^\T$.}
  \label{fig:attnS-cnn}
\end{figure}

The $H$-head linear attention with separate rank-one key and query can be viewed as a sum of $H$ three-layer convolutional linear network with the cubic feature $\vz$ defined in \cref{eq:def-feat-z} as input. Specifically, \cref{eq:def-attnS-rank1} can be rewritten as 
\begin{align}
\attn_{\text S}(\mX)_{D+1,N+1} = \sum_{i=1}^H v_i \vq_i^\T \mK_i \vz
,\nonumber \\ \text{where } \,
\mK_i = \begin{bmatrix}
\vk_i^\T & \vzero_D^\T & \hdots & \vzero_D^\T \\
\vzero_D^\T & \vk_i^\T & \hdots & \vzero_D^\T  \\
\vdots & \vdots & \ddots & \vdots  \\
\vzero_D^\T & \vzero_D^\T & \hdots & \vk_i^\T
\end{bmatrix}
\in \sR^{D\times D^2}  .
\end{align}
The matrix $\mK_i$ is a convolutional matrix with kernel size $D$ and stride $D$.
A schematic of the three-layer convolutional linear network is given in \cref{fig:attnS-cnn}.

When the number of heads satisfies $H\geq D$, the linear attention with separate rank-one key and query, $\attn_{\text S}(\mX)$, can express any linear map of $\vz(\mX)$ and has the same expressivity as linear attention with merged key and query, $\attn_{\text M}(\mX)$. However, the two models correspond to multi-layer linear networks with different connectivity and depths, resulting in different loss landscape \citep{kohn22geometry,kohn24function} and training dynamics \citep{saxe14exact,saxe19semantic}.

\begin{figure*}
\vspace{-1ex}
  \centering
  \subfloat[Loss  \label{fig:attnS-loss}]
    {\centering
    \includegraphics[width=0.3\linewidth]{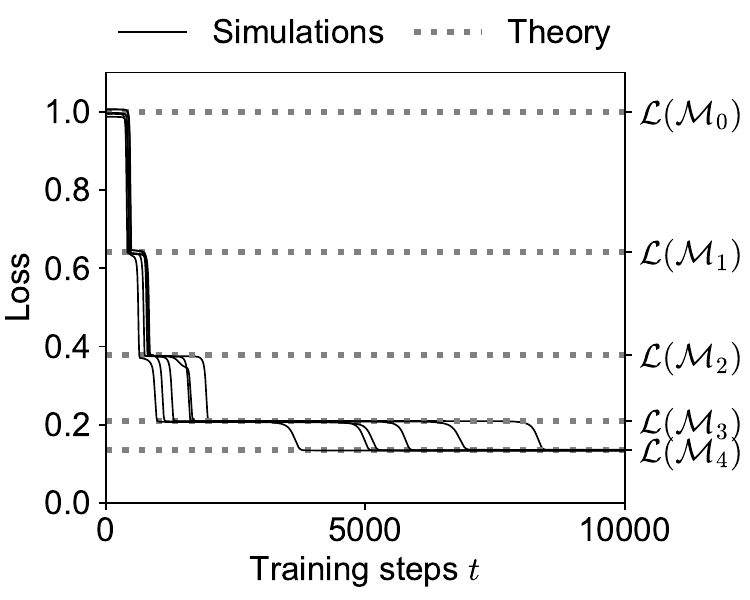}}
    \hfill
  \subfloat[Loss and value weights  \label{fig:attnS-value}]
    {\centering
    \includegraphics[width=0.255\linewidth]{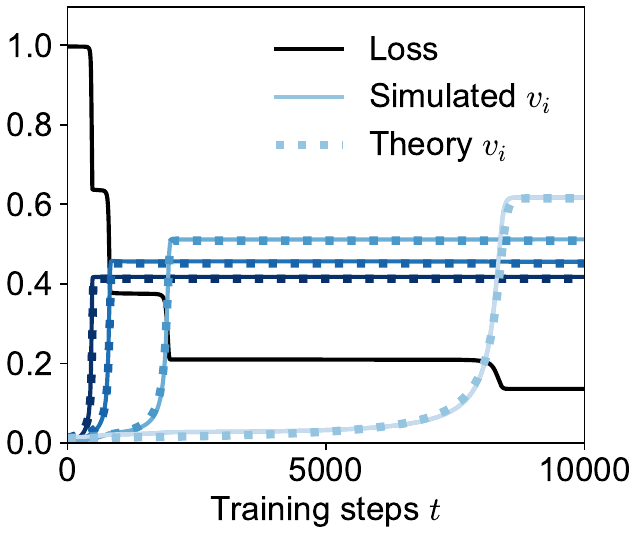}}
  \subfloat[Key weights and eigenvectors of $\mLambda$  \label{fig:attnS-W}]
    {\centering
    \includegraphics[width=0.44\linewidth]{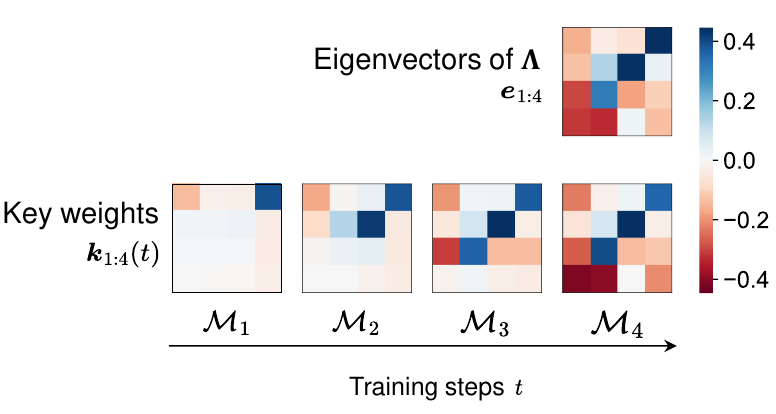}}
  \vspace{-1ex}
  \caption{Multi-head linear attention with separate rank-one key and query exhibits saddle-to-saddle dynamics.
  (a) The loss curve has $D$ abrupt drops, separated by plateaus (six runs from different random initialization are plotted). The loss at each plateau matches our theoretical prediction in \cref{eq:attnS-loss} (dashed gray lines). 
  (b) The value weight $v_i$ in each head for one of the runs in (a) is plotted in solid blue curves. The numerical solutions of $v_i$ from \cref{eq:1Ddynamics} are plotted in dashed blue curves and match the simulations well. The shades of blue distinguish different heads.
  (c) The key weights during the loss plateau are plotted in color. When the model moves from one fixed point to the next, the key weight in a head, $\vk_i$, aligns with a new eigenvector of the input token covariance $\mLambda$. The key weights $\vk_{1:4}$ and the eigenvectors $\ve_{1:4}$ are rows in the heatmaps.
  A video of the dynamics is provided at \href{https://yedizhang.github.io/img/attnS.gif}{URL}.
  Here $D=4,N=31,H=4$, and $\mLambda$ has eigenvalues $0.4,0.3,0.2,0.1$ and eigenvectors as plotted in (c).}
  \label{fig:attnS}
\end{figure*}

\subsection{Loss Landscape: Exponentially Many Fixed Points  \label{sec:attnS-landscape}}
The gradient flow training dynamics of linear attention with separate rank-one key and query, derived in \cref{supp:attnS-gd}, is given by
{\allowdisplaybreaks
\begin{subequations}  \label{eq:attnS-gd}
\begin{align}
\tau \dot v_i &=  \vk_i^\T \left( \mLambda^2 - \E\left({\hat\mLambda}^2\right) \sum_{i'=1}^H v_{i'} \vk_{i'} \vq_{i'}^\T \mLambda \right) \vq_i  ,\\
\tau \dot \vk_i &= v_i \left( \mLambda^2 - \E\left({\hat\mLambda}^2\right) \sum_{i'=1}^H v_{i'} \vk_{i'} \vq_{i'}^\T \mLambda \right) \vq_i  ,\\
\tau \dot \vq_i &= v_i \left( \mLambda^2 - \mLambda \sum_{i'=1}^H v_{i'} \vk_{i'} \vq_{i'}^\T \E\left({\hat\mLambda}^2\right) \right) \vk_i  ,
\end{align}
\end{subequations}}%
where we denote the in-context covariance of $\vx_n$ tokens as $\hat\mLambda = \sum_{n=1}^N \vx_n \vx_n^\T / N$ and the expectation of ${\hat\mLambda}^2$ is
\begin{align}
\E\left({\hat\mLambda}^2\right) = \mLambda^2 + \frac{\mLambda+\tr(\mLambda)\mI}N \mLambda
\end{align}
This dynamical system contains $2^D$ fixed points in the function space of $\attn_{\text S} (\mX)_{D+1,N+1}$. We specify the fixed points below and prove their validity in \cref{supp:attnS-landscape}. 

Let $\lambda_1,\cdots,\lambda_D$ be the eigenvalues of the covariance matrix $\mLambda$ arranged in descending order, and $\ve_1,\cdots,\ve_D$ be the corresponding normalized eigenvectors.
We use $\gM(\gS_m)$ to denote a set of fixed points that correspond to learning $m$ ($m=0,1,\cdots,D$) out of the $D$ eigenvectors,
% \begin{align}  \label{eq:attnS-def-M}
% \gM(\gS_m)= \left\{ v_{1:H}, \vk_{1:H}, \vq_{1:H} \big| \text{conditions (C1)-(C3)} \right\}  ,
% \end{align}
\begin{align}  \label{eq:attnS-def-M}
\gM(\gS_m)= \left\{ (v, \vk, \vq)_{1:H} \big| \text{conditions (C1)-(C3)} \right\}  ,
\end{align}
where the set $\gS_m$ specifies the indices of the learned eigenvectors,
\begin{align}  \label{eq:def-Sm}
\gS_m \subseteq \{1,2,\cdots,D\} ,\, | \gS_m | = m .
\end{align}
The three conditions for \cref{eq:attnS-def-M} are:
\vspace{-1ex}
\begin{enumerate}
    \item[(C1)] The heads sum to fit the eigenvectors with indices in the set $\gS_m$
    \vspace{-4ex}
\end{enumerate}
\begin{align}  \label{eq:attnS-condition1}
    \sum_{i=1}^H v_i \vk_i \vq_i^\T = \sum_{d\in \gS_m} \lambda_d^{-1} \left( 1 + \frac{1+\tr(\mLambda)/\lambda_d}N \right)^{-1} \ve_d \ve_d^\T.
\end{align}
\begin{enumerate}[topsep=0pt]
    \item[(C2)] For heads with a nonzero value weight, $v_i\neq0$, both $\vk_i$ and $\vq_i$ lie in the span of $\{ \ve_d \}_{d\in \gS_m}$.
    \item[(C3)] For heads with a zero value weight, $v_i=0$, at least one of $\vk_i$ or $\vq_i$ lies in the span of $\{ \ve_d \}_{d\in \gS_m}$.
\end{enumerate}
Since there are ${D \choose m}$ possible ways of choosing $m$ out of $D$ indices to define $\gS_m$ in \cref{eq:def-Sm}, the total number of possible choices summed over $m=0,\cdots,D$ is $\sum_{m=0}^D {D \choose m} = 2^D$. Each choice corresponds to a different condition (C1) in \cref{eq:attnS-condition1} and thus a different function, $\attn_{\text S} (\mX)_{D+1,N+1}$. Hence, the gradient flow dynamics in \cref{eq:attnS-gd} has $2^D$ fixed points in the function space.
\footnote{A fixed point in function space corresponds to a set of fixed points in weight space that implement the same input-output map.}

% In comparison with \cref{sec:attnM-landscape} where there were two fixed points in the dynamics of linear attention with merged key and query, there are now $2^D$ fixed points for linear attention with separate key and query. 
The two fixed points of $\attn_{\text M}$ (\cref{sec:attnM-landscape}) are contained in the $2^D$ fixed points of $\attn_{\text S}$: the zero fixed point in \cref{eq:attnM-M0} corresponds to $\gM(\gS_0)$, i.e., learning no eigenvector; the global minimum fixed point in \cref{eq:attnM-Mstar} corresponds to $\gM(\gS_D)$, i.e., learning all $D$ eigenvectors.

\subsection{Training Dynamics: Saddle-to-Saddle Dynamics  \label{sec:attnS-dynamics}}
Building on the exponentially many fixed points we have identified, we now analyze which fixed points are actually visited in gradient flow training and in what order. We find that starting from small initialization, the model visits $(D+1)$ out of the $2^D$ fixed points. 

With small initialization, the model is initially near the unstable zero fixed point, $\gM_0=\gM(\varnothing)$. As training progresses, the model sequentially visits the fixed points in $\gM_1,\gM_2,\cdots,\gM_D$, where $\gM_m = \gM(\{ 1,2,\cdots,m \})$.
That is, the model trained from small initialization sequentially learns to fit the first eigenvector (the eigenvector of $\mLambda$ with the largest eigenvalue), the second eigenvector, and so on. As shown in \cref{fig:attnS-loss}, the loss goes through $D$ abrupt drops in training, each corresponding to the transition from one fixed point to the next. The abrupt drops of loss are separated by plateaus, during which the model lingers near an unstable fixed point. Because the time required for a head to learn the eigenvector $\ve_m$ from small initialization scales with $\lambda_m^{-2}$ (see \cref{supp:M1-M2}), eigenvectors associated with larger eigenvalues are learned faster. This explains why the model learns to fit the eigenvectors sequentially in descending order of the eigenvalues, as well as why we empirically see the later plateaus last longer in \cref{fig:attnS-loss}.

When the model is at a fixed point in $\gM_m$, we compute the loss in \cref{supp:attnS-loss} and obtain
\begin{align}  \label{eq:attnS-loss}
\Ls(\gM_m) = \tr(\mLambda) - \sum_{d=1}^m \lambda_d \left( 1 + \frac{1+\tr(\mLambda)/\lambda_d}N \right)^{-1}  .
\end{align}
\cref{eq:attnS-loss} is highly interpretable in the limit of a large sequence length $N$. The loss, $\Ls(\gM_m)$, is the sum of the eigenvalues associated with the remaining unlearned eigenvectors
\begin{align*}
\lim_{N\to \infty}
\Ls(\gM_m) = \tr(\mLambda) - \sum_{d=1}^m \lambda_d = \sum_{d=m+1}^D \lambda_d  .
\end{align*}
Thus, the loss decreases by approximately $\lambda_m$ during the $m$-th abrupt loss drop. We plot \cref{eq:attnS-loss} as dashed gray lines in \cref{fig:attnS-loss} and find they match the plateaus of simulated loss trajectories well.

When the model reaches $\gM_m$ from small initialization, its weights take on a highly structured form, which is a specific instance of the general definition in \cref{eq:attnS-def-M}. As shown in \cref{fig:attnS-W}, the key and query weights in a head grow in scale and align with a new eigenvector of the input token covariance $\mLambda$ during each abrupt loss drop. Based on simulations in \cref{fig:attnS} and derivations in \cref{supp:M0-M1,supp:M1-M2}, we propose an ansatz that during the $(m+1)$-th plateau ($0\leq m < D$) and the subsequent abrupt drop of loss, the weights are approximately given by\footnote{We permute the heads so that the head aligned with the $d$-th eigenvector have index $d$. The signs of any two among $v_i,\vk_i,\vq_i$ can be flipped with trivial effect on the analysis.}
\begin{subequations}  \label{eq:attnS-Mm-ansatz}
\begin{align}
\vk_i &= \vq_i = v_i \ve_i ,\,
v_i = \lambda_i^{-\frac13} \left( 1 + \frac{1+\tr(\mLambda)/\lambda_i}N \right)^{-\frac13} ,  \nonumber\\
&\hspace{20.5ex} 1 \leq i \leq m ,  \\
\vk_i &= \vq_i = v_i(t) \ve_{m+1} ,\quad
i = m+1 ,  \\
\vk_i &= \vq_i = \vzero, v_i=0 ,\quad
m+2 \leq i \leq H ,
\end{align}
\end{subequations}
where $v_{m+1}(t)$ is small during the $(m+1)$-th loss plateau and grows during the $(m+1)$-th abrupt loss drop.
\cref{eq:attnS-Mm-ansatz} implies that the $\ell^2$ norms of $v_i,\vk_i,\vq_i$ in a head are equal, which is a consequence of small initialization and the conservation law in \cref{supp:attnS-conservation}.
With this ansatz, the high-dimensional training dynamics during the $(m+1)$-th plateau and the subsequent abrupt drop of loss reduces to an ordinary differential equation about $v_i(t), i=m+1$:
\begin{align}  \label{eq:1Ddynamics}
\tau \dot v_i = \lambda_{m+1}^2 v_i^2 - \lambda_{m+1}^3 \left( 1 + \frac{1+\tr(\mLambda)/\lambda_{m+1}}N \right) v_i^5 .
\end{align}
\cref{eq:1Ddynamics} is a separable differential equation but does not admit a general analytical solution of $v_{m+1}(t)$ in terms of $t$ (see \cref{eq:wolfram-solution}). Nonetheless, it greatly simplifies the high-dimensional dynamics in \cref{eq:attnS-gd} and provides a good approximation of the true dynamics: during each plateau and the subsequent abrupt loss drop, weights in one of the heads grow in scale with the key and query weights aligning with the next eigenvector, while the rest of the heads remain approximately unchanged. In \cref{fig:attnS-value}, we compare the numerical solution of \cref{eq:1Ddynamics} with the value weights trajectories in the simulation and find excellent agreement.

In summary, the loss trajectory of linear attention with separate rank-one key and query trained from small initialization exhibits $D$ abrupt drops, each followed by a plateau. The amount of the $m$-th abrupt loss drop ($1 \leq m \leq D$) is approximately the eigenvalue $\lambda_m$, during which the key and query weights in an attention head grow in scale and align with the eigenvector $\ve_m$.

\subsection{ICL Algorithm: Principal Component Regression}
When the linear attention model is at a fixed point in $\gM_m$, based on \cref{eq:attnS-condition1}, the model implements 
\begin{align}  \label{eq:attnS-PCR}
&\attn_{\text S} (\mX)_{D+1,N+1}  \nonumber\\
=& \vbeta^\T \sum_{d=1}^m \lambda_d^{-1} \left( 1 + \frac{1+\tr(\mLambda)/\lambda_d}N \right)^{-1} \ve_d \ve_d^\T \vx_q .
\end{align}
In the limit of a large sequence length $N$, \cref{eq:attnS-PCR} simplifies and can be interpreted as principal component regression in context with $m$ principal components
\begin{align*}
\lim_{N\to \infty} \attn_{\text S} (\mX)_{D+1,N+1} = \vw^\T \sum_{d=1}^m \ve_d \ve_d^\T \vx_q  .
\end{align*}
Here $\vw$ is the task vector for the sequence $\mX$, and $\sum_{d=1}^m \ve_d \ve_d^\T \vx_q$ is query input $\vx_q$ projected onto the first $m$ principal components. Hence, if training stops during the $(m+1)$-th plateau, the linear attention approximately implements the principal component regression algorithm in context with $m$ principal components.

After the model has undergone $D$ plateaus, it converges to the global minimum fixed point, $\gM_D$, and approximately implements principal component regression in context with all $D$ components, which is least square regression. Thus, the linear attention model with either merged or separate key and query undergoes different training dynamics but converges to the same global minimum solution.

\begin{figure*}
  \centering
  \includegraphics[width=\linewidth]{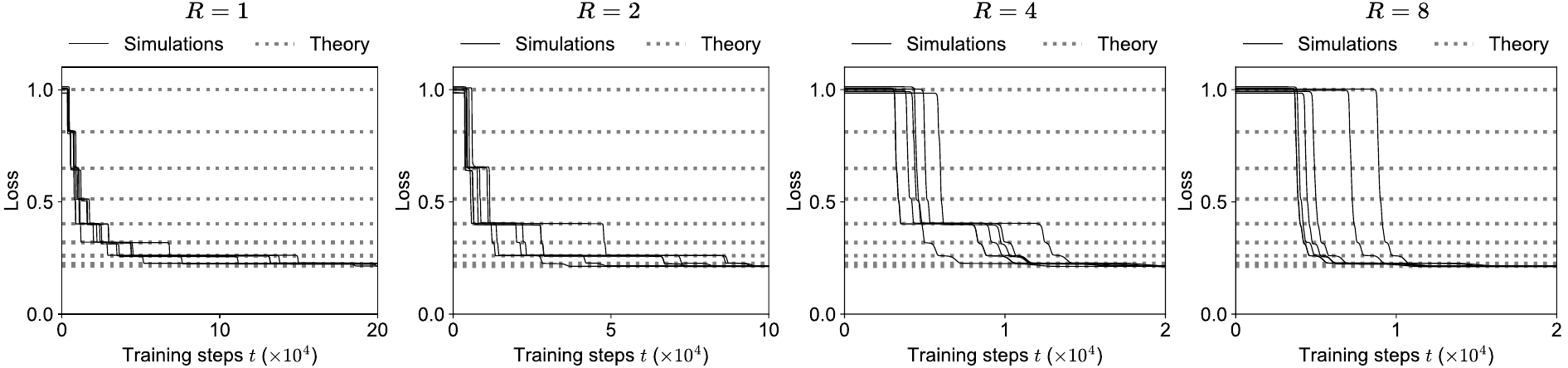}
  \vspace{-4.4ex}
  \caption{Multi-head linear attention with separate low-rank key and query exhibits saddle-to-saddle dynamics, with the duration of plateaus depending on the rank $R$. Solid black curves are loss trajectories from six random initializations. Dashed gray lines mark the loss values predicted by \cref{eq:attnS-loss} at nine fixed points, which are $\Ls(\gM_0),\Ls(\gM_1),\cdots,\Ls(\gM_8)$ from top to bottom. The four panels differ only in the rank of the key and query weights. Here $D=8,N=31,H=9$, $\mLambda$ has trace $1$ and eigenvalues $\lambda_d\propto d^{-1}$.}
  \label{fig:attnS-lowrank}
\end{figure*}

\section{Linear Attention with Separate Low-Rank Key and Query  \label{sec:attnS-lowrank}}
The linear attention model with separate rank-$R$ key and query shares many behaviors with its rank-one counterpart. 
For loss landscape, linear attention with rank-$R$ key and query has the same $2^D$ fixed points in the function space as its rank-one counterpart, corresponding to the model implementing in-context principal component regression with a subset of all $D$ principal components (see \cref{supp:attnS-lowrank-landscape}).

For training dynamics, the loss trajectories differ slightly, depending on the rank $R$. 
We plot the loss trajectories with input token dimension $D=8$ and different ranks $R=1,2,4,8$ in \cref{fig:attnS-lowrank} (see \cref{fig:attnS-lowrank-supp} for $R=3,5,6,7$).
For $R=1$, the loss exhibits plateaus at eight values $\Ls(\gM_m) \, (m=0,1,\cdots,7)$.
For $R=2$, the loss exhibits plateaus at four values $\Ls(\gM_m) \, (m=0,2,4,6)$, and either brief plateaus or no plateau at the other four values.
For $R=4$, the loss exhibits conspicuous plateaus at only two values $\Ls(\gM_m) \, (m=0,4)$.
To summarize, with rank-$R$ key and query, the loss trajectory exhibits conspicuous plateaus at value $\Ls(\gM_m)$ for $m$ that divides $R$.

The difference in the loss trajectories arises from the structure of the model defined in \cref{eq:def-attnS}. Each attention head has a single value weight $v_i$ that is associated with all $R$ pairs of key and query weights in that head, $\vk_{i,r}, \vq_{i,r} \, (r=1,\cdots,R)$. During a conspicuous plateau, a new value weight escapes from the unstable zero fixed point and grows in scale. Once the value weight has grown, it leads to larger gradient updates for all the key and query weights in that head, speeding up their escape from the zero fixed point.
Hence, in the rank-$R$ case, a conspicuous plateau occurs when $m$ divides $R$, corresponding to learning a new head from small initialization. Brief or no plateau occurs when $m$ does not divide $R$, corresponding to learning a new pair of key and query weights in a head whose value weight has already grown, as shown in \cref{fig:attnS-lowrank-value}.
See \cref{supp:attnS-lowrank-explain-s2s} for further details.

\section{Related Work}
Recent theoretical research on linear attention has investigated its expressivity \citep{oswald24versatile,gatmiry24looping}, learnability \citep{yau24polytime}, loss landscape \citep{mahankali24linear,li24finegrain}, convergence \citep{zhang24jmlr,zhang24mlp,ren24bigram,fu24convergence}, and generalization \citep{wu24howmany,mahankali24linear,duraisamy24finite,cengiz24asymptotic,belkin24scaling,frei24benign}. 
The seminal work by \citet{zhang24jmlr} analyzed the gradient flow training dynamics of linear attention to prove convergence guarantees, showing what the model converges to at the end of training. Our work also analyzes the gradient flow training dynamics but goes beyond existing convergence results to describe the entire training dynamics. Moreover, we study multi-head attention with merged or separate key and query weights, while \citet{zhang24jmlr} focused on single-head attention with merged key and query.

% Another line of recent research on the training dynamics of softmax attention models has shown stage-wise dynamics across different settings, including phenomena such as the increasing rank of weights \citep{boix23rank} and learning higher-order interactions among input tokens \citep{rende24simplicity,edelman24markov}. However, due to the intractability of softmax attention training dynamics in general, many of these studies made certain assumptions to enable theoretical analyses, including restricted weights \citep{boix23rank,siyu24dynamics,rende24simplicity,edelman24markov}, specifically chosen datasets \citep{huang24convergence}, and a simpler layer-wise training algorithm in place of standard gradient descent \citep{tian23snap,nichani24casual,siyu24ih,wang24ih}.
Another line of recent research on the training dynamics of softmax attention models has shown stage-wise dynamics. Due to the intractability of softmax attention training dynamics in general, many of these studies made strong assumptions to enable theoretical analyses, including a simplified layer-wise training algorithm in place of standard gradient descent \citep{tian23snap,nichani24casual,siyu24ih,wang24ih}, restricted weights \citep{boix23rank,siyu24dynamics,rende24simplicity,edelman24markov}, and specifically chosen datasets \citep{huang24convergence}.
In comparison, our work leverages the linear attention model without the softmax operation, enabling us to study in fine detail the dynamics of standard gradient descent training without restrictions on weights. 
Namely, we derive an analytical time-course solution and reduce the high-dimensional dynamics to one-dimensional ordinary differential equations for the two models we study, respectively.
Furthermore, we characterize how parametrization (i.e., merged or separate key and query, and rank of the separate key and query weights) affects the loss landscape and training dynamics, an aspect not previously examined.

\section{Discussion}
We studied the gradient flow training dynamics of multi-head linear attention and demonstrated how it acquires ICL abilities in training.
We begin with a simple setting of linear attention with merged key and query trained for in-context linear regression, following the setting in seminal works \citep{oswald23GD,ahn23preconditioned,zhang24jmlr}. We show an abrupt loss drop in training and give an analytical time-course solution in the case of a white input token covariance and small initialization. 
However, a single abrupt loss drop does not fully capture the evolution of ICL in training practical transformers, where the abilities continue to develop throughout training \citep{xia23trajectory,park24competition}. We thus extend our analysis to a parametrization closer to the attention in practical transformers: attention with separate key and query. In the separate case, we find that the loss exhibits saddle-to-saddle dynamics with multiple abrupt drops. The ICL ability evolves progressively, manifesting as implementing principal component regression in context, with the number of principal components increasing over training time.
% Building on prior findings showing that transformers can implement different forms of ICL \citep{bai23statistician,lampinen24spectrum}, we show that different forms of ICL can indeed emerge in gradient descent training. 
%By identifying the in-context algorithm at different times in training, {\color{red} not sure what ``at different times in training'' means here} we characterize how
We thus characterize how the linear attention model develops increasingly sophisticated ICL abilities in gradient descent training.

\begin{figure}
  \centering
    \subfloat[Softmax $\attn_{\text M}$]
    {\centering
    \includegraphics[width=0.495\linewidth]{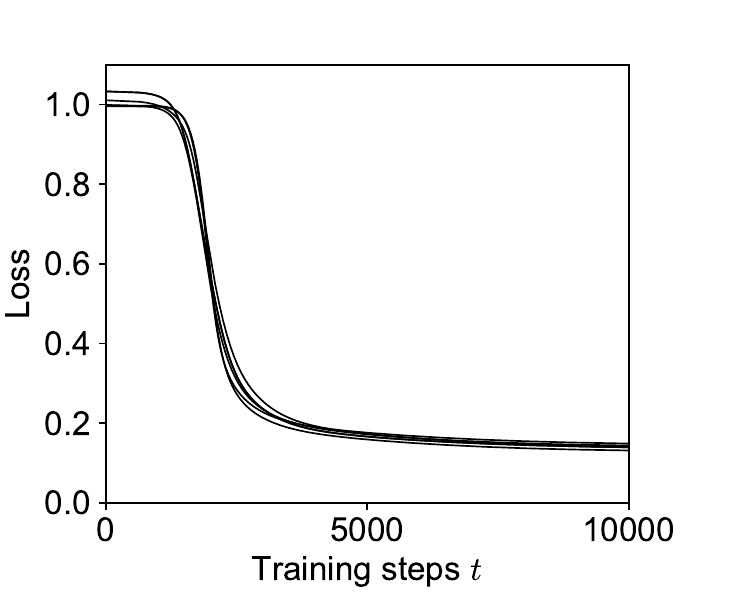}}
    \subfloat[Softmax $\attn_{\text S}$ with $R=1$]
    {\centering
    \includegraphics[width=0.5\linewidth]{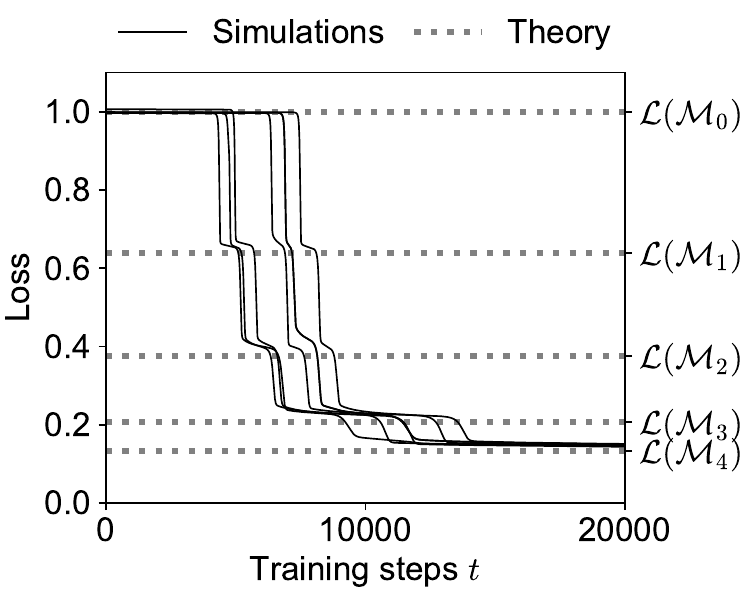}}
  \caption{Loss trajectories of softmax attention with merged or separate key and query. Six runs from different random initialization are plotted. Similar to the linear attention case, softmax $\attn_{\text M}$ exhibits one abrupt loss drop, while softmax $\attn_{\text S}$ exhibits multiple loss drops. The dataset and model setup are the same as \cref{fig:attnM,fig:attnS}, except for adding the softmax activation function.}
  \label{fig:softmax-attn}
\end{figure}

\textbf{Softmax Attention}.
We empirically find that the different training dynamics of linear $\attn_{\text M}$ and linear $\attn_{\text S}$ also occur in their softmax counterparts. \cref{fig:softmax-attn} follows the same setup as \cref{fig:attnM,fig:attnS} for linear attention, with the only difference being adding the softmax activation function for the attention calculation. We observe that softmax attention with merged key and query exhibits a single abrupt loss drop, whereas softmax attention with separate rank-one key and query undergoes multiple loss drops, separated by phases of conspicuously slower training. This suggests that our findings and theoretical intuition are not unique to linear attention but may also extend to softmax attention.

\begin{figure}
\vspace{1.4ex}
  \centering
    \subfloat[Linear $\attn_{\text M}$  \label{fig:init-sweep-attnM}]
    {\centering
    \includegraphics[width=0.48\linewidth]{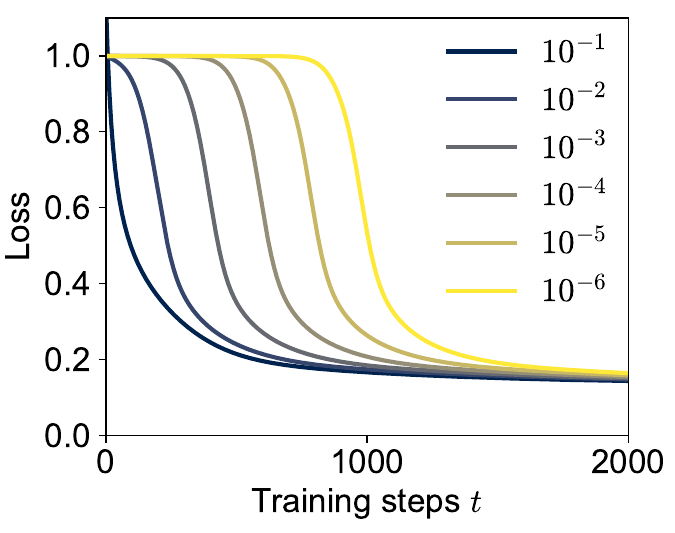}}
\hfill
    \subfloat[Linear $\attn_{\text S}$ with $R=1$  \label{fig:init-sweep-attnS}]
    {\centering
    \includegraphics[width=0.48\linewidth]{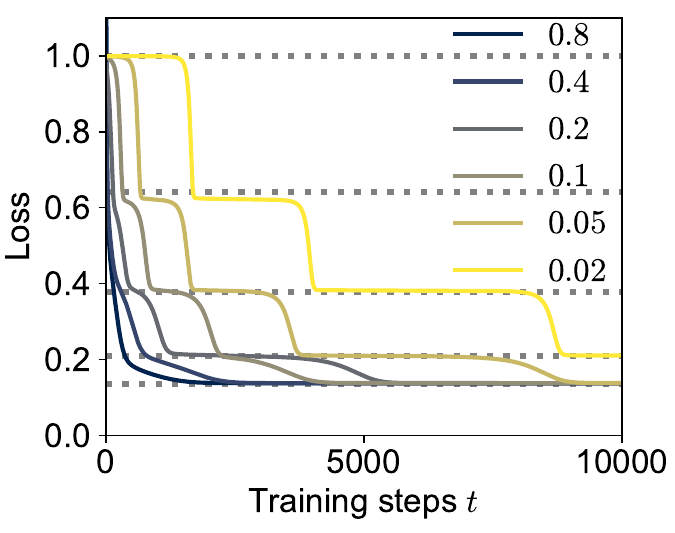}}
  \caption{Loss trajectories of linear $\attn_{\text M}$ and $\attn_{\text S}$ with varying initialization scales. The colors indicate the initialization scale. Increasing the initialization scale shortens the plateaus. The dataset and model setup are the same as \cref{fig:attnM,fig:attnS}, except that we vary the initialization scale.}
  \label{fig:init-sweep}
\end{figure}

\textbf{Effect of Initialization}.
Having analyzed the small initialization case, we now examine how the initialization scale affects training dynamics.
For linear $\attn_{\text M}$, \cref{fig:init-sweep-attnM} shows increasing initialization shortens the plateau before the single abrupt loss drop. For linear $\attn_{\text S}$, \cref{fig:init-sweep-attnS} shows increasing initialization shortens all plateaus between successive abrupt loss drops. 
At the largest initialization, both models exhibit an exponential-shaped loss decay -- a hallmark of lazy learning \citep{chizat19lazy}.
In contrast, rich learning typically exhibits abrupt sigmoid-shaped loss curves as seen in our main result.
Theory typically focuses on either the lazy or rich regime, while practical initializations often fall in between.
In \cref{fig:init-sweep}, dynamics from the intermediate initialization seems like a mix of the exponential-shaped and sigmoid-shaped curves, which are often seen in practice, e.g. in induction head emergence in natural language settings \citep[Argument 1]{anthropic22ih}. 
Our analysis focuses on the rich regime and takes a first step toward understanding dynamics in naturalistic settings.

\textbf{Dynamics of In-Context and In-Weight Learning}.
Our work studies the training dynamics of in-context learning. Other than in-context learning, attention models can also learn in weight; that is, solving the task by memorizing the map between the query input and the target output without using the information in context. The arbitration between in-context and in-weight learning may depend on properties of training data \citep{chan22icliwl}.
To focus on the dynamics of ICL, we used a purely ICL task, which is in-context linear regression with the task vector sampled from a zero-mean standard normal distribution, $\vw \sim \mathcal N(\vzero,\mI)$. Since memorizing any particular task vector does not effectively decrease the loss, linear attention develops only in-context learning ability during training, as shown in \cref{fig:icl-iwl-p0}.
If the task vector $\vw$ follows a different distribution, the training dynamics involves the development of both in-context and in-weight learning abilities, as shown in \cref{fig:icl-iwl-sweep}. We provide more details in \cref{supp:icl-iwl}.

\textbf{Implications for Future Theory}.
% \citep{baldi89pca,fukumizu98batch,saxe14exact,saxe19semantic,arora18acc,ji18align,atanasov22silent}
In our analysis, we draw connections between linear attention and multi-layer linear networks, enabling us to employ the rich theoretical machinery built for linear networks to understand linear attention training dynamics. Beyond training dynamics, many other theoretical results for linear networks can apply to linear attention through the equivalence we draw. For example, the convergence guarantee for multi-head linear attention trained on in-context linear regression tasks can be obtained from the convergence proofs for deep linear networks \citep{arora19convergence,shamir19convergence}. In contrast, without the equivalence, \citet{zhang24jmlr} previously obtained a convergence guarantee for single-head linear attention, which required highly non-trivial derivations. Hence, we believe the connections we draw are useful in enabling the applications of theory from one architecture to the other.

Additionally, we have shown that parametrization significantly affects the loss landscape and training dynamics, motivating future research to examine how their results may or may not be influenced by the parametrization choice.

\section*{Reproducibility}
Code reproducing our main results is available at GitHub:
\texttt{
\href{https://github.com/yedizhang/linattn-icl}{https://github.com/yedizhang/linattn-icl}}

\section*{Acknowledgement}
We thank Jin Hwa Lee, Sara Dragutinovi\'c, Andrew Lampinen, Basile Confavreux and William Tong for helpful conversations.

We thank the following funding sources: Gatsby Charitable Foundation (GAT3850) to YZ, AKS, PEL, and AS; Wellcome Trust (110114/Z/15/Z) to PEL; Sainsbury Wellcome Centre Core Grant from Wellcome (219627/Z/19/Z) to AS; Schmidt Science Polymath Award to AS. AS is a CIFAR Azrieli Global Scholar in the Learning in Machines \& Brains program.

\section*{Impact Statement}
This paper presents work whose goal is to advance the field of machine learning. There are many potential societal consequences of our work, none which we feel must be specifically highlighted here.

\newpage
\bibliography{ref}
\bibliographystyle{icml2025}

\newpage
\appendix
\onecolumn
\input{supp}

\end{document}

%% file: math_cmd.tex
\usepackage{amsmath, amssymb, amsthm, amsfonts, bm, bbm}

\newcommand{\T}{\top}
\newcommand{\attn}{\mathsf{ATTN}}
\newcommand{\VEC}{\texttt{vec}}
\newcommand{\tr}{\texttt{tr}}
\newcommand{\init}{\text{init}}
\newcommand{\diff}{\mathrm{d}}
\def\eqref#1{equation~\ref{#1}}
\def\1{\bm{1}}

\def\vzero{{\bm{0}}}

\def\vbeta{{\bm{\beta}}}

\def\va{{\bm{a}}}
\def\vb{{\bm{b}}}

\def\ve{{\bm{e}}}

\def\vk{{\bm{k}}}

\def\vm{{\bm{m}}}

\def\vq{{\bm{q}}}

\def\vu{{\bm{u}}}
\def\vv{{\bm{v}}}
\def\vw{{\bm{w}}}
\def\vx{{\bm{x}}}

\def\vz{{\bm{z}}}

\def\mA{{\bm{A}}}
\def\mB{{\bm{B}}}
\def\mC{{\bm{C}}}
\def\mD{{\bm{D}}}

\def\mI{{\bm{I}}}

\def\mK{{\bm{K}}}

\def\mM{{\bm{M}}}

\def\mU{{\bm{U}}}

\def\mW{{\bm{W}}}
\def\mX{{\bm{X}}}

\def\mLambda{{\bm{\Lambda}}}

\def\mOmega{{\bm{\Omega}}}

\DeclareMathAlphabet{\mathsfit}{\encodingdefault}{\sfdefault}{m}{sl}
\SetMathAlphabet{\mathsfit}{bold}{\encodingdefault}{\sfdefault}{bx}{n}

\def\gM{{\mathcal{M}}}

\def\gS{{\mathcal{S}}}

\def\sR{{\mathbb{R}}}

\newcommand{\E}{\mathbb{E}}
\newcommand{\Ls}{\mathcal{L}}

%% file: supp.tex
\setcounter{tocdepth}{2}
\renewcommand{\contentsname}{\centering Table of Contents}
{\hypersetup{linkcolor=black}\small 
\baselineskip=0.8\baselineskip
\setlength{\parskip}{0.8pt}
 \tableofcontents}

\newpage
\bigskip
\centerline{\textbf {\large Appendix}}

\section{Additional Figures \label{supp:add-figs}}
\subsection{Varying Context Lengths  \label{supp:varylen}}
\begin{figure}[h]
  \centering
    \subfloat[Linear $\attn_{\text M}$]
    {\centering
    \includegraphics[width=0.33\linewidth]{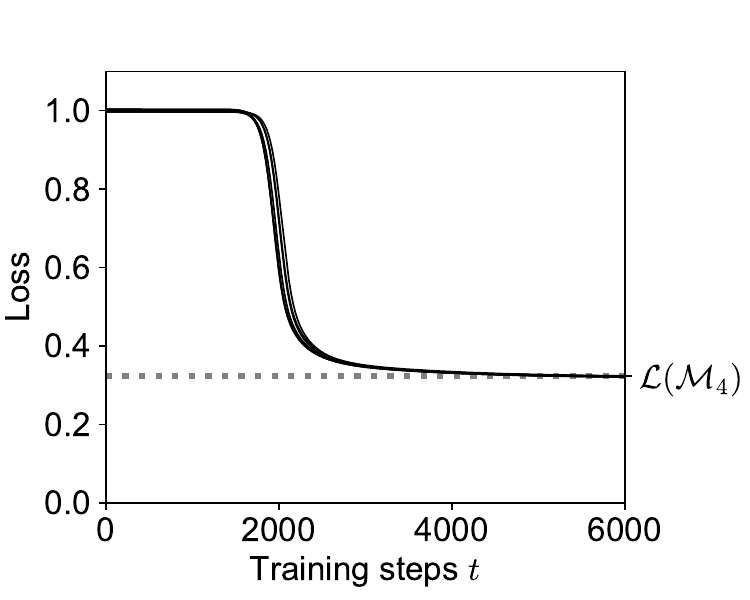}}
    \hspace{2ex}
    \subfloat[Linear $\attn_{\text S}$ with $R=1$]
    {\centering
    \includegraphics[width=0.33\linewidth]{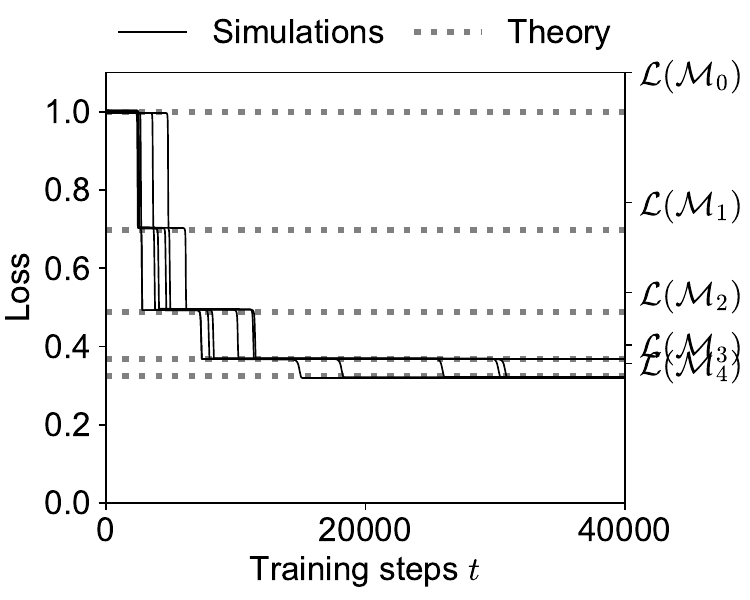}}
  \caption{Loss trajectories of linear $\attn_{\text M}$ and $\attn_{\text S}$ trained with the next token prediction loss defined in \cref{eq:next-token-loss} with $N_{\mathrm {max}}=31$. In this case, the models are trained on sequences of varying lengths, which they can handle due to the $1/N$ scaling factor in \cref{eq:def-attnM,eq:def-attnS,eq:def-beta}. The dataset and model setup are the same as \cref{fig:attnM,fig:attnS}, except that we switch to the next token prediction loss.}
  \label{fig:varylen}
\end{figure}

In our main results, we consider a fixed context length $N$, because our training sequences have the same length and the loss is computed only for the last query token, $\Ls = \E (y_q - \hat y_q)^2$. In practice, however, the training sequences may have varying lengths, and the loss can be computed for every token in the sequence, that is
\begin{align}  \label{eq:next-token-loss}
\Ls_{\text{ntp}} = \E \left[ \frac1{N_{\mathrm {max}}} \sum_{n=2}^{N_{\mathrm {max}}+1} (y_n - \hat y_n)^2 \right] ,
\end{align}
where $y_{N+1} = y_q$, and $\hat y_n$ is the attention model's prediction for $y_n$ when given only the first $n$ columns of $\mX$ as input. 
We demonstrate how our results apply to the case of varying context lengths. Specifically, the distribution of the context lengths only influences our results through a statistic, $\mathbb E(1/N)$.

For $\mathsf{ATTN_M}$, derivations in \cref{supp:attnM-converged} show that the converged model implements
\begin{align}  \label{eq:attnM-converged-varylen-raw}
\attn_{\text M} (\mX)_{D+1,N+1} = \sum_{i=1}^H v_i \vbeta^\T \mU_i \vx_q
= \vbeta^\top \left[ \mathbb E \left( \frac1N \vx_n \vx_n^\top  \right)^2 \right]^{-1} \mLambda \vx_q .
\end{align}
Substituting \cref{eq:ELambdaLambda-varylen} into \cref{eq:attnM-converged-varylen-raw}, we obtain
\begin{align}  \label{eq:attnM-converged-varylen}
\attn_{\text M} (\mX)_{D+1,N+1} = \vbeta^\T \left[ \mLambda + \mathbb E \left(\frac1N\right) (\mLambda + \tr(\mLambda) \mI) \right]^{-1} \vx_q  .
\end{align}
The distribution of context lengths only influences \cref{eq:attnM-converged-varylen} through the expectation $\mathbb E(1/N)$. For a fixed context length, $\mathbb E (1/N)=1/N$, which recovers \cref{eq:attnM-converged} in the main text. For the next token prediction loss, the distribution of context lengths, $p(N)$, follows a uniform distribution over $\{1,2,\cdots, N_{\mathrm {max}}\}$. The expectation $\mathbb E(1/N)$ is the harmonic number divided by $N_{\mathrm {max}}$, which doesn't have a closed-form expression but can be easily computed for a specific finite $N_{\mathrm {max}}$.

Similarly, for $\attn_{\text S}$ trained with varying context lengths, the fixed point condition (C1) takes the form
\begin{align*}
\sum_{i=1}^H v_i \vk_i \vq_i^\T = \sum_{d\in \gS_m} \lambda_d^{-1} \left[ 1 + E\left(\frac1N\right) (1+\tr(\mLambda)/\lambda_d) \right]^{-1} \ve_d \ve_d^\T ,
\end{align*}
where the expectation $\mathbb E(1/N)$ reduces to $1/N$ in the fixed context length case as in \cref{eq:attnS-condition1}. Consequently, when the model is at a fixed point in $\gM_m$, the loss value is
\begin{align}  \label{eq:attnS-loss-varylen}
\Ls(\gM_m) = \tr(\mLambda) - \sum_{d=1}^m \lambda_d \left[ 1 + E\left(\frac1N\right) (1+ \tr(\mLambda)/\lambda_d) \right]^{-1}  ,
\end{align}
which reduces to \cref{eq:attnS-loss} when $\mathbb E(1/N)=1/N$.

We train $\attn_{\text M}$ and $\attn_{\text S}$ with the next token prediction loss as in \cref{eq:next-token-loss} and plot the loss trajectories in \cref{fig:varylen}. The loss trajectories are qualitatively similar to those in \cref{fig:attnM,fig:attnS-loss}, modulo the different loss values during the plateaus. We plot the loss values computed from \cref{eq:attnS-loss-varylen} as dashed gray lines and find they match the plateaus of the simulated loss trajectories well.

\subsection{Higher Dimensions}
We train $\attn_{\text M}$ and $\attn_{\text S}$ on a dataset with larger $N,D$. The loss trajectories are qualitatively similar to those in lower-dimensional cases, despite being noisier. This suggests that our findings do not break in high-dimensional settings.
\begin{figure}[h]
  \centering
    \subfloat[Linear $\attn_{\text M}$]
    {\centering
    \includegraphics[width=0.25\linewidth]{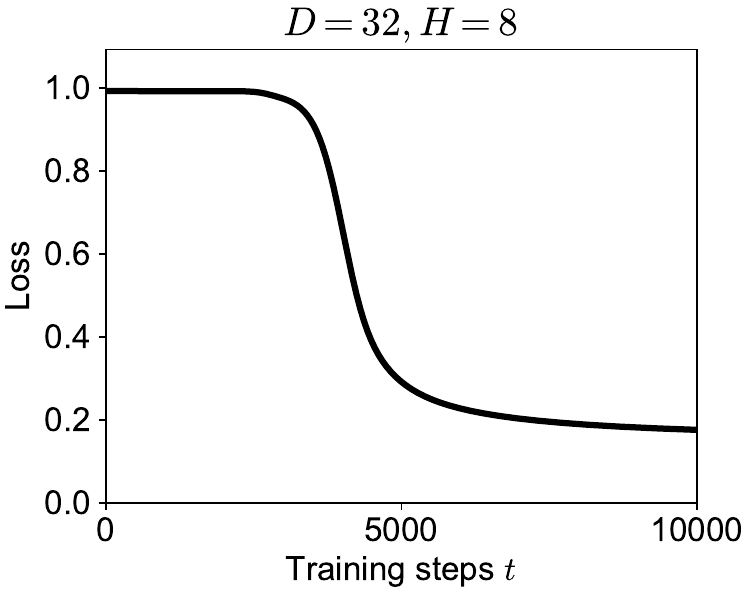}
    \includegraphics[width=0.25\linewidth]{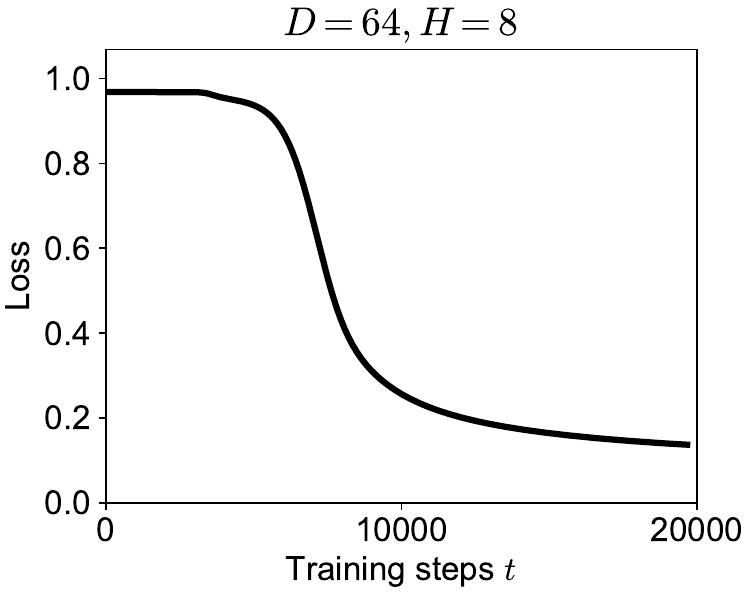}}
    \subfloat[Linear $\attn_{\text S}$]
    {\centering
    \includegraphics[width=0.25\linewidth]{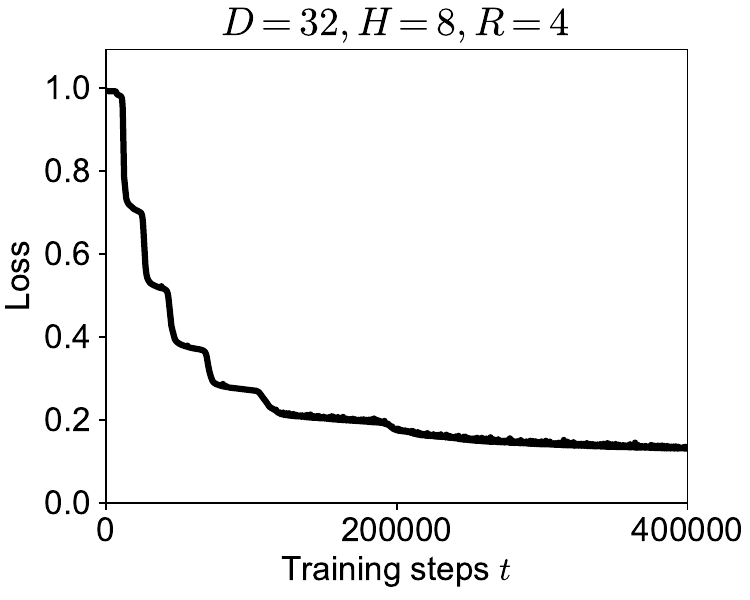}
    \includegraphics[width=0.25\linewidth]{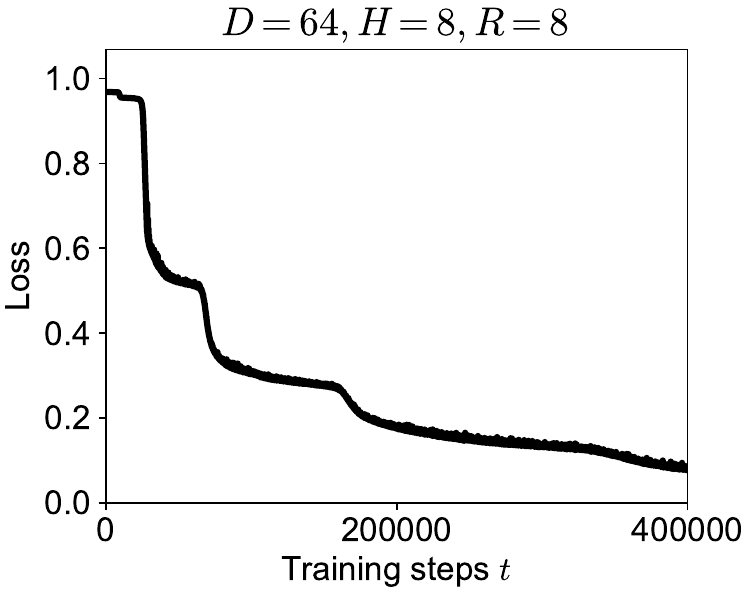}}
  \caption{Loss trajectories of linear $\attn_{\text M}$ and $\attn_{\text S}$ with high-dimensional data. Here the sequence length is $N=127$, $\mLambda$ has trace $1$ and eigenvalues $\lambda_d\propto d^{-1}$. Other hyperparameters are labeled at the top of each panel.}
\end{figure}

\section{Additional Related Work}
A concurrent work by \citet{geshkovski24metastability} studies saddle-to-saddle-like dynamics in softmax attention models following a mathematical framework for slow motion of gradient flows \citep{otto07slowmotion}. A subsequent work by \citet{he25softmax} examines the training dynamics of softmax attention trained on the in-context linear regression task; that is, the case we briefly touch on in \cref{fig:softmax-attn}.

A broader body of theoretical literature have explored the transformers training dynamics but addressed different problem from ours, such as the effect of initialization \citep{makkuva24init}, convergence results \citep{song24dynamics,rhuang24nonasymptotic}, sample complexity guarantees \citep{ildiz24markov}, scaling limits \citep{bordelon24infinite}, and implicit regularization \citep{ataee23maxmargin,tarzanagh24svm,julistiono24mirror,vasudeva24implicit,sheen24reg}. Other studies considered special training regimes, such as the neural tangent kernel regime \citep{jang24loraNTK} and the mean-field regime \citep{kim24meanfield}. A few works focused on vision transformers \citep{jelassi22vit,jiang24vit,huang24vit}. In contrast, our work focuses on the training dynamics and the development of ICL abilities over time.

It is recognized that transformers can perform ICL, whereas it is an open question whether fully-connected networks can perform ICL \citep{lee24iclarch,boix24abstract,tong24mlp}. In \cref{sec:attnM-mlp}, we revealed an equivalence between linear attention and a fully-connected linear network with cubic feature input, which is an instance of a fully-connected network performing ICL. Furthermore, we demonstrate that fully-connected networks may perform ICL more comparably to attention models when provided with polynomial features instead of the original sequence. This may explain why \citet[Figure 25]{boix24abstract} observed that fully-connected networks fail to learn ICL with the original sequence as input, but succeed when the input is augmented with $\mX \mX^\T$.

\section{Additional Preliminaries}
\subsection{Data Statistics}
Recall that we use $\vbeta$ to denote the in-context correlation between $\vx_n$ and $y_n$ in a sequence $\mX$, as defined in \cref{eq:def-beta}. We additionally denote the in-context covariance of $\vx_n$ in a sequence as $\hat \mLambda$
\begin{align}
\hat \mLambda \equiv \frac1N \sum_{n=1}^N \vx_n \vx_n^\T .
\end{align}
We can thus write $\mX\mX^\T/N$ as a block matrix
\begin{align}
\frac1N \mX\mX^\T 
= \begin{bmatrix}
\frac1N \left(\vx_q \vx_q^\T + \sum_n \vx_n \vx_n^\T \right) & \frac1N \sum_n \vx_n y_n  \\
\frac1N \sum_n y_n \vx^\T & \frac1N \sum_n y_n^2
\end{bmatrix}
= \begin{bmatrix}
\frac1N \vx_q \vx_q^\T + \hat \mLambda & \vbeta  \\
\vbeta^\T & \vw^\T \hat\mLambda \vw
\end{bmatrix} .
\end{align}
Due to the definition of the in-context linear regression task, we have that 
\begin{align}
\vbeta = \hat \mLambda \vw .
\end{align}
We will need a statistic, $\E\left({\hat\mLambda}^2\right)$. Let $p(N)$ denote the distribution of context lengths, and recall that $\vx_n \sim \mathcal N(\vzero, \mLambda)$. We obtain:
\begin{align}  \label{eq:ELambdaLambda-varylen}
\E \left({\hat\mLambda}^2\right) 
&\equiv \E \left( \frac1N \sum_{n=1}^N \vx_n \vx_n^\T \right)^2  \nonumber\\
&= \E \left( \frac{N^2-N}{N^2} \sum_{n\neq n'} \vx_n \vx_n^\T \vx_{n'} \vx_{n'}^\T + \frac{N}{N^2}\sum_{n=1}^N \vx_n \vx_n^\T \vx_n \vx_n^\T \right)  \nonumber\\
&= \E_N \left( \frac{N-1}N \right) \E_\vx \left( \vx_n \vx_n^\T \right) \E_\vx \left( \vx_{n'} \vx_{n'}^\T \right) + \E_N \left( \frac1N \right) \E_\vx \left( \vx_n \vx_n^\T \vx_n \vx_n^\T \right)   \nonumber\\
&= \left( 1 - \E \left( \frac1N \right) \right) \mLambda^2 + \E \left( \frac1N \right) \left( 2\mLambda^2 + \tr(\mLambda) \mLambda \right)  \nonumber\\
&= \mLambda^2 + \E \left( \frac1N \right) \left( \mLambda+\tr(\mLambda)\mI \right) \mLambda  .
\end{align}
For our main results, we use a fixed context length, that is $p(N)$ is a point mass distribution and $\E (1/N)=1/N$. In this case, \cref{eq:ELambdaLambda-varylen} simplifies to
\begin{align}  \label{eq:ELambdaLambda}
\E\left({\hat\mLambda}^2\right) = \mLambda^2 + \frac{\mLambda+\tr(\mLambda)\mI}N \mLambda  .
\end{align}
We note that the eigenvectors of $\E\left({\hat\mLambda}^2\right) $ are the same as those of $\mLambda$, which are $\ve_1,\cdots,\ve_D$,
\begin{align*}
\E\left({\hat\mLambda}^2\right) \ve_d
= \left( 1+\frac1N \right) \mLambda^2 \ve_d + \frac{\tr(\mLambda)}N \mLambda \ve_d
= \left[ \left( 1+\frac1N \right) \lambda_d^2 + \frac{\tr(\mLambda)}N \lambda_d \right] \ve_d .
\end{align*}
We denote the eigenvalues of $\E({\hat\mLambda}^2)$ corresponding to eigenvectors $\ve_1,\cdots,\ve_D$ as $a_1,\cdots,a_D$. These eigenvalues are given by
\begin{align}  \label{eq:def-ad}
a_d = \left[ \left( 1+\frac1N \right) \lambda_d^2 + \frac{\tr(\mLambda)}N \lambda_d \right]
= \lambda_d^2 \left( 1 + \frac{1+\tr(\mLambda)/\lambda_d}N \right) .
\end{align}
The matrix $\E({\hat\mLambda}^2)$ can be expressed through its eigen-decomposition, which will be useful in later derivations:
\begin{align}  \label{eq:LambdaLambda-eig}
\E\left({\hat\mLambda}^2\right) = \sum_{d=1}^D a_d \ve_d \ve_d^\T .
\end{align}

\subsection{Initialization}
For linear attention with merged key and query, we initialize the entries of the value and the merged key-query weights as
\begin{align}
v_i \sim \mathcal N (0,w_\init^2/H)  ,\quad
U_i^{d,d'} \sim \mathcal N (0,w_\init^2/HD^2)  .
\end{align}
At initialization, the following $\ell^2$ norms are 
\begin{align}
\sqrt{\sum_{i=1}^H v_i^2}, \sqrt{\sum_{i=1}^H \| \mU_i\|^2}  \sim O(w_\init) .
\end{align}
For linear attention with separate rank-$R$ key and query, we initialize the entries of the value, key, and query weights as
\begin{align}
v_i \sim \mathcal N (0,w_\init^2/H)  ,\quad
k_{i,r}^d \sim \mathcal N (0,w_\init^2/HRD)  ,\quad
q_{i,r}^d \sim \mathcal N (0,w_\init^2/HRD)  .
\end{align}
At initialization, the following $\ell^2$ norms are 
\begin{align}
\sqrt{\sum_{i=1}^H v_i^2}, \sqrt{\sum_{i=1}^H\sum_{r=1}^R \| \vk_{i,r}\|^2}, \sqrt{\sum_{i=1}^H\sum_{r=1}^R \| \vq_{i,r}\|^2}  \sim O(w_\init) .
\end{align}

\subsection{Kronecker Product  \label{supp:kronecker}}
The Kronecker product, denoted as $\otimes$, is defined for two matrices of arbitrary sizes. The Kronecker product of the matrix $\mA \in \sR^{p\times q}$ and the matrix $\mB \in \sR^{r\times s}$ is a block matrix of shape $pr \times qs$ 
\begin{align*}
\mA \otimes \mB = \begin{bmatrix}
a_{11} & \cdots & a_{1q}  \\
\vdots & \ddots & \vdots  \\
a_{p1} & \cdots & a_{pq}
\end{bmatrix} \otimes \mB
= \begin{bmatrix}
a_{11}\mB & \cdots & a_{1q}\mB  \\
\vdots & \ddots & \vdots  \\
a_{p1}\mB & \cdots & a_{pq}\mB
\end{bmatrix}  .
\end{align*}
Based on the definition, it holds for any pair of column vectors $\va$ and $\vb$
\begin{align}
\va \otimes \vb = \VEC(\vb \va^\T)  .  \label{eq:kronecker-vec}
\end{align}
We quote some properties of the Kronecker product to be used in our derivations:
\begin{subequations}  \label{eq:kronecker}
\begin{align}
(c \mA) \otimes \mB &= \mA \otimes (c \mB) = c (\mA \otimes \mB)
\quad\text{for any scalar } c, \\
(\mA \otimes \mB)^\T &= \mA^\T \otimes \mB^\T
\quad\text{for any matrices } \mA,\mB,  \\
(\mA \otimes \mB)^{-1} &= \mA^{-1} \otimes \mB^{-1}
\quad\text{for invertible matrices } \mA,\mB,  \label{eq:kronecker-inverse} \\
(\mA \otimes \mB) (\mC \otimes \mD) &= (\mA \mC) \otimes (\mB \mD)
\quad\text{for compatible matrices } \mA,\mB,\mC,\mD,  \\
(\mB^\T \otimes \mA) \VEC(\mM) &= \VEC (\mA \mM \mB)
\quad\text{for compatible matrices } \mA,\mB,\mM .  \label{eq:kronecker-mix}
\end{align}
\end{subequations}

\section{Linear Attention with Merged Key and Query}
\subsection{Justification for Zero Blocks Assumption  \label{supp:attnM-zerouv}}
We prove our claim in \cref{sec:attnM-def} that $\vv_i$ and $\vu_i$  remain zero throughout training if their initialization is zero.

\begin{proof}
The bottom right entry of $\attn_{\text M}(\mX)$ is given by
\begin{align*}  
\hat y_q \equiv \attn_{\text M} (\mX)_{D+1,N+1}
&= \sum_{i=1}^H \begin{bmatrix}
\vv_i^\T & v_i
\end{bmatrix}
\begin{bmatrix}
\frac1N \left(\vx_q \vx_q^\T + \sum_n \vx_n \vx_n^\T \right) & \frac1N \sum_n \vx_n y_n  \\
\frac1N \sum_n y_n \vx_n^\T & \frac1N \sum_n y_n^2
\end{bmatrix}
\begin{bmatrix}
\mU_i \\ \vu_i^\T
\end{bmatrix} \begin{bmatrix}
\vx_q \\ 0
\end{bmatrix}
\\
&= \sum_{i=1}^H \left( \vv_i^\T \left( \hat \mLambda + \frac1N \vx_q \vx_q^\T \right) \mU_i + v_i \vbeta^\T \mU_i + \vv_i^\T \vbeta \vu_i^\T + v_i \vw^\T \hat \mLambda \vw \vu_i^\T \right) \vx_q  .
\end{align*}
If we initialize $\vv_i, \vu_i=\vzero$, $\hat y_q$ is
\begin{align*}
\hat y_q = \sum_{i=1}^H v_i \vbeta^\T \mU_i \vx_q = \vw^\T \hat \mLambda \sum_{i=1}^H v_i \mU_i \vx_q .
\end{align*}
We now calculate the gradient updates of $\vv_i, \vu_i$ and prove their gradients are zero if their initialization is zero. 
The gradient update of $\vv_i$ contains $\E(\vw)$, which is zero. Specifically, we have, from \cref{eq:grad-flow},
\begin{align}  \label{eq:attnM-zerov}
\tau \dot \vv_i &= \E \left[ (y_q - \hat y_q) \left( \left( \hat \mLambda + \frac1N \vx_q \vx_q^\T \right) \mU_i + \vbeta \vu_i^\T \right) \vx_q \right]  \nonumber\\
&= \E \left[ \left( \vw^\T \vx_q - \vw^\T \hat\mLambda \sum_{i=1}^H v_i \mU_i \vx_q \right) \left( \hat \mLambda + \frac1N \vx_q \vx_q^\T \right) \mU_i \vx_q \right]  \nonumber\\
&= \E_\vw (\vw)^\T \E \left[ \left( \vx_q - \hat\mLambda \sum_{i=1}^H v_i \mU_i \vx_q \right) \left( \hat \mLambda + \frac1N \vx_q \vx_q^\T \right) \mU_i \vx_q \right]  \nonumber\\
&= \vzero .
\end{align}
Note that we separated the expectation of $\vw$ because of the independence between $\vw$ and all $\vx$ tokens.

The gradient update of $\vv_i$ contains $\E_\vw \left( \vw^\T \hat \mLambda \vw \vw^\T \right)$, whose entries are linear combinations of third moments of the zero-mean normal random variable $\vw$, and are thus zero. Specifically, we have
\begin{align}  \label{eq:attnM-zerou}
\tau \dot \vu_i &= \E \left[ \left( \vv_i^\T \vbeta + v_i \vw^\T \hat\mLambda \vw \right) (y_q - \hat y_q) \vx_q \right]  \nonumber\\
&= \E \left[ v_i \vw^\T \hat \mLambda \vw \left( \vw^\T \vx_q - \vw^\T \hat \mLambda \sum_{i=1}^H v_i \mU_i \vx_q \right) \vx_q \right]  \nonumber\\
&= \E_\vw \left( \vw^\T \hat \mLambda \vw \vw^\T \right) \E \left[ v_i \left( \vx_q - \hat \mLambda \sum_{i=1}^H v_i \mU_i \vx_q \right) \vx_q \right]  \nonumber\\
&= \vzero .
\end{align}
\end{proof}

\subsection{Gradient Flow Equations}
We here derive the gradient flow dynamics for linear attention with merged key and query given in \cref{eq:attnM-gd}.

For linear attention with merged key and query, the prediction for the query output can be written as $\hat y_q = \vw_2^\T \mW_1 \vz$ due to 
\cref{eq:attnM-mlp}. Based on the gradient flow training rule in \cref{eq:grad-flow}, the gradient flow dynamics is
\begin{align*}
\tau \dot \mW_1 &= \E \left[ \vw_2 \left(y_q - \vw_2^\T \mW_1 \vz \right) \vz^\T \right]
= \vw_2 \left( \E \left(y_q \vz^\T \right) - \vw_2^\T \mW_1 \E \left(\vz \vz^\T \right) \right) ,   \\
\tau \dot \vw_2 &= \E \left[ \mW_1 \left(y_q - \vw_2^\T \mW_1 \vz \right) \vz \right]
= \mW_1 \left( \E \left(y_q \vz^\T \right) - \vw_2^\T \mW_1 \E \left(\vz \vz^\T \right) \right)^\T ,
\end{align*}
which was introduced in \cref{eq:attnM-gd} in the main text.

\subsection{Fixed Points}
To find the fixed points, we set the gradients in \cref{eq:attnM-gd} to zero
\begin{align*}
\tau \dot \mW_1 &= \vw_2 \left( \E \left(y_q \vz^\T \right) - \vw_2^\T \mW_1 \E \left(\vz \vz^\T \right) \right) \overset{\text{set}}{=} \vzero
,   \\
\tau \dot \vw_2 &= \mW_1 \left( \E \left(y_q \vz^\T \right) - \vw_2^\T \mW_1 \E \left(\vz \vz^\T \right) \right)^\T \overset{\text{set}}{=} \vzero  ,
\end{align*}
which yield the two manifolds of fixed points introduced in \cref{eq:attnM-def-M} in the main text:
\begin{align*}
\vw_2 = \vzero, \, \mW_1 = \vzero 
\quad &\Rightarrow \quad
\gM_0 = \{ \vw_2 = \vzero,  \mW_1 = \vzero \}   , \\
\E \left(y_q \vz^\T \right) - \vw_2^\T \mW_1 \E \left(\vz \vz^\T \right) = \vzero
\quad &\Rightarrow \quad
\gM_* = \left\{ \vw_2, \mW_1 \big| \vw_2^\T \mW_1 = \E \left( y_q \vz^\T \right) \E \left( \vz \vz^\T \right)^{-1} \right\}  .
\end{align*}

\subsection{Duality of the Global Minimum Solution  \label{supp:attnM-converged}}
We here prove the second equality in \cref{eq:attnM-converged}, that is
\begin{align*} 
\E \left( y_q \vz^\T \right) \E \left( \vz \vz^\T \right)^{-1} \vz
= \vbeta^\T \left( \mLambda + \frac{\mLambda + \tr(\mLambda) \mI}N \right)^{-1} \vx_q  .
\end{align*}
This equality implies the intriguing duality that the linear regression solution in the cubic feature space of $\vz$ is the in-context linear regression solution in the original space of the $\vx_n,y_n$ token pairs for each sequence $\mX$.
\begin{proof}
We first calculate the input and input-output correlations in the cubic feature space. We denote $\mLambda_q \equiv \E\left( \vx_q \vx_q^\T \right)$. While $\mLambda_q =\mLambda$, this equality is not needed in this proof.

Due to the property in \cref{eq:kronecker-vec}, the cubic feature $\vz$ can be written as 
\begin{align}
    \vz=\VEC \left( \vbeta \vx_q^\T  \right)= \vx_q \otimes \vbeta  .
\end{align}
We substitute in $y_q=\vx_q^\T \vw,\, \vz=\vx_q \otimes \vbeta$ and use the properties of the Kronecker product in \cref{eq:kronecker} to obtain
\begin{align}  \label{eq:Eyz}
\E \left( y_q \vz^\T \right)
&= \E \left[ \vx_q^\T \vw (\vx_q^\T \otimes \vbeta^\T) \right]  \nonumber\\
&= \E \left( \vx_q^\T \otimes \vx_q^\T \vw \vw^\T \hat \mLambda \right)  \nonumber\\
&= \E \left( \vx_q^\T \otimes \vx_q^\T \hat \mLambda \right)  \nonumber\\
&= \E \VEC \left( \hat \mLambda \vx_q \vx_q^\T \right)^\T  \nonumber\\
&= \VEC (\mLambda \mLambda_q)^\T  .
\end{align} 
Similarly, we have
\begin{align}  \label{eq:Ezz}
\E \left( \vz \vz^\T \right)
&= \E \left[ ( \vx_q \otimes \vbeta) (\vx_q^\T \otimes \vbeta^\T) \right]  \nonumber\\
&= \E \left[ (\vx_q \vx_q^\T) \otimes (\vbeta \vbeta^\T) \right]  \nonumber\\
&= \E(\vx_q \vx_q^\T) \otimes \E(\hat \mLambda \vw \vw^\T \hat \mLambda)  \nonumber\\
&= \mLambda_q \otimes \E\left( {\hat\mLambda}^2 \right)  .
\end{align}
Using \cref{eq:kronecker-inverse}, the inverse of $\E \left( \vz \vz^\T \right)$ is
\begin{align}  \label{eq:Ezz-inv}
\E \left( \vz \vz^\T \right)^{-1} = \mLambda_q^{-1} \otimes \E\left( {\hat\mLambda}^2 \right)^{-1} .
\end{align}
Multiplying \cref{eq:Eyz,eq:Ezz-inv} with $\vz=\vx_q \otimes \vbeta$, and using \cref{eq:kronecker-mix} twice, we obtain
\begin{align}
\E \left( y_q \vz^\T \right) \E \left( \vz \vz^\T \right)^{-1} \vz
&= \VEC (\mLambda \mLambda_q)^\T \mLambda_q^{-1} \otimes \E\left( {\hat\mLambda}^2 \right)^{-1} (\vx_q \otimes \vbeta)  \nonumber\\
&= \VEC \left[ \E\left( {\hat\mLambda}^2 \right)^{-1} \mLambda \mLambda_q \mLambda_q^{-1} \right]^\T  (\vx_q \otimes \vbeta)  \nonumber\\
&= \vbeta^\T \E\left( {\hat\mLambda}^2 \right)^{-1} \mLambda \vx_q .
\end{align}
Substituting in $\E\left( {\hat\mLambda}^2 \right)$ obtained from \cref{eq:ELambdaLambda} finishes the proof
\begin{align*}
\E \left( y_q \vz^\T \right) \E \left( \vz \vz^\T \right)^{-1} \vz
= \vbeta^\T \left( \mLambda^2 + \frac{\mLambda+\tr(\mLambda)\mI}N \mLambda \right)^{-1} \mLambda \vx_q
= \vbeta^\T \left( \mLambda + \frac{\mLambda+\tr(\mLambda)\mI}N \right)^{-1} \vx_q .
\end{align*}
\end{proof}

\subsection{Analytical Time-Course Solution for White Covariance  \label{supp:attnM-solution}}
We include a derivation of the time-course solution of two-layer fully-connected linear network with white input covariance and vanishing initialization following \cite{saxe14exact}, and then apply it to linear attention. 
With vanishing initialization, the conserved quantity given in \cref{eq:attnM-conservation} is exactly zero throughout learning,
\begin{align*}
\vw_2 \vw_2^\T - \mW_1 \mW_1^\T = \vzero .
\end{align*}
Hence, there exists a unit norm vector $\vm$ such that $\mW_1 = \vw_2 \vm^\T$. With the assumption of white covariance, $\E \left( \vz \vz^\T \right)=\alpha\mI_{D^2}$, \cite{saxe14exact,atanasov22silent} have shown that the unit norm vector $\vm$ is parallel with the correlation between $y_q$ and $\vz$ throughout training, that is
\begin{align}  \label{eq:change-variable}
\mW_1 = \vw_2 \vm^\T
,\quad \text{where } \vm = \frac{\E ( y_q \vz )}{\|\E ( y_q \vz )\|} .
\end{align}
We substitute \cref{eq:change-variable} and the white covariance assumption, $\E \left( \vz\vz^\T\right)=\alpha\mI_{D^2}$, into the gradient flow dynamics given in \cref{eq:attnM-gd} and obtain
\begin{align*}
\tau \dot \vw_2 = \vw_2 \vm^\T \left( \E ( y_q \vz ) - \alpha \vw_2^\T \vw_2 \vm  \right)
= \vw_2 \left( \gamma - \alpha  \vw_2^\T \vw_2 \right)
,\quad \text{where } \gamma \equiv \|\E ( y_q \vz )\| .
\end{align*}
Notice that the square of the $\ell^2$ norm of $\vw_2$ follows a solvable ordinary differential equation. Let $s=\vw_2^\T \vw_2$. The dynamics of $s(t)$ is
\begin{align}  \label{eq:ode-s}
\tau \dot s = 2 \vw_2^\T \tau \dot \vw_2 = 2 \vw_2^\T \vw_2 \left( \gamma - \alpha  \vw_2^\T \vw_2 \right) = 2 s (\gamma - \alpha s)  .
\end{align}
We can solve this differential equation by separating variables and integrating both sides,
\begin{align*}
\int_{s(0)}^{s(t)} \frac{1}{s (\gamma - \alpha s)} \diff s = \int_0^t \frac{2}{\tau} \diff t  
\quad \Rightarrow \quad
\frac1\gamma \ln \frac{s(t)(\gamma-\alpha s(0))}{s(0)(\gamma-\alpha s(t))} = \frac{2}{\tau}t  .
\end{align*}
The solution of $s(t)$ is given by
\begin{align*}  
s(t) &= \frac{\gamma e^{2\gamma\frac t\tau}}{\alpha \left(e^{2\gamma\frac t\tau}-1\right)+\frac{\gamma}{s(0)}} .
\end{align*}
The time-course of the total weights is given by
\begin{align}  \label{eq:saxe-solution}
\vw_2^\T \mW_1 = s(t) \vm^\T = s(t) \frac{\E ( y_q \vz )^\T}{\|\E ( y_q \vz )^\T\|} .
\end{align}
We now apply this solution to linear attention. If the input token covariance is identity, $\mLambda=\mI_D$, we calculate the input and input-output correlations in the cubic feature space according to \cref{eq:Eyz,eq:Ezz} and get
\begin{align*}
\E \left( y_q \vz^\T \right) &= \VEC(\mI_D)^\T  , \\
\E \left( \vz \vz^\T \right) &= \mI_D \otimes \left( 1+\frac{1+D}N \right) \mI_D
= \left( 1+\frac{1+D}N \right) \mI_{D^2} .
\end{align*}
The parameters in the dynamics of the equivalent two-layer fully-connected linear network are
\begin{align}  \label{eq:alpha-gamma}
\alpha = 1+\frac{1+D}N   ,\, 
\gamma = \left\| \VEC(\mI_D) \right\| = \sqrt D  .
\end{align}
Substituting \cref{eq:alpha-gamma} into \cref{eq:saxe-solution}, we obtain 
\begin{align*}
\vw^\T \mW_1 (t) = s(t) \frac{\VEC(\mI_D)^\T}{\sqrt D}
,\quad \text{where } 
s(t) =  \frac{\sqrt D e^{2\sqrt D\frac t\tau}}{\left( 1+\frac{1+D}N \right) \left(e^{2\sqrt D\frac t\tau}-1\right)+\frac{\sqrt D}{s(0)}}  .
\end{align*}
Due to the equivalence between linear attention and the two-layer fully-connected linear network given in \cref{eq:attnM-mlp}, we obtain
\begin{align*}
\attn_{\text M} (\mX;t)_{D+1,N+1} 
= \vw_2^\T \mW_1 \vz 
= s(t) \frac{\VEC(\mI_D)^\T}{\sqrt D} \vx_q \otimes \vbeta 
= \frac1{\sqrt D} s(t) \vbeta^\T \mI_D \vx_q
= \frac1{\sqrt D} s(t) \vbeta^\T \vx_q .
\end{align*}
which is \cref{eq:attnM-solution} in the main text where we have rewritten $\sigma(t) = s(t) / \sqrt D$.

The time-course of loss can also be expressed in terms of $\sigma(t)$ as
\begin{align}  \label{eq:attnM-solution-ls}
\Ls(t) = \left( 1 - 2\sigma (t) + \left( 1+\frac{1+D}N \right) \sigma(t)^2 \right) D .
\end{align}

\subsection{Training Dynamics for General Covariance}
\subsubsection{Early Dynamics Predicts Duration of Plateau  \label{supp:attnM-duration}}
For a general input covariance matrix, the full time-course solution to two-layer fully-connected linear networks is currently unavailable. Nonetheless, the training dynamics is well understood and we can analyze the early phase dynamics to estimate the duration of the loss plateau.

In the early phase of training when the loss plateaus, the weights have not moved much away from their small initialization. The training dynamics of $\mW_1$ is mainly driven by the first term in \cref{eq:attnM-gd-W1}, and similarly for $\vw_2$ in \cref{eq:attnM-gd-w2}
\begin{align*}
\tau \dot \mW_1 &= \vw_2 \left( \E \left(y_q \vz^\T \right) - \vw_2^\T \mW_1 \E \left(\vz \vz^\T \right) \right) 
= \vw_2 \E \left(y_q \vz^\T \right) + O(w_\init^3)  , \\
\tau \dot \vw_2 &= \mW_1 \left( \E \left(y_q \vz^\T \right) - \vw_2^\T \mW_1 \E \left(\vz \vz^\T \right) \right)^\T
= \mW_1 \E \left(y_q \vz \right) + O(w_\init^3)  .
\end{align*}
Thus, the early training dynamics is well approximated by the linear dynamical system
\begin{align*}
\tau \dot \mW_1 = \vw_2 \E \left(y_q \vz^\T \right)  ,\quad
\tau \dot \vw_2 = \mW_1 \E \left(y_q \vz \right)  .
\end{align*}
In the case of nonwhite covariance, the change of variable in \cref{eq:change-variable} is valid in the early phase of training but no longer valid when the loss starts to decrease appreciably \citep{atanasov22silent}. For the early training dynamics, we apply the change of variable in \cref{eq:change-variable} and obtain
\begin{align*}
\tau \dot \vw_2 = \vw_2 \vm^\T \E \left(y_q \vz \right) = \gamma \vw_2 .
\end{align*}
Recall that $s=\vw_2^\T \vw_2$. The early phase dynamics of $s(t)$ is approximately
\begin{align*}
\tau \dot s = 2 \gamma s .
\end{align*}
We solve the differential equation and obtain 
\begin{align*}
t =  \frac{\tau}{2\gamma} ( \ln s(t) - \ln s(0) )  .
\end{align*}
Due to small initialization, $\ln s(t)$ at the end of the plateau is much smaller compared to $-\ln s(0)$. Hence, the duration of the initial plateau of loss, $t_{\text{plateau}}$, is
\begin{align}  \label{eq:attnM-duration}
t_{\text{plateau}} \approx \frac{\tau}{2\gamma} \ln \frac1{s(0)}  .
\end{align}
Here, the scalar $\gamma$ is
\begin{align}  \label{eq:attnM-gamma}
\gamma \equiv \|\E ( y_q \vz )\| = \left\| \E \left( \vw^\T \vx_q \VEC(\vbeta \vx_q^\T) \right) \right\|
&= \left\| \E \left( \vw^\T \vx_q \vbeta \vx_q^\T \right) \right\|_{\text F}  \nonumber\\
&= \left\| \E \left( \hat\mLambda \vw \vw^\T \vx_q \vx_q^\T \right) \right\|_{\text F}   \nonumber\\
&= \left\| \E \left( \hat\mLambda \right) \E_\vw \left( \vw \vw^\T \right) \E_{\vx_q} \left( \vx_q \vx_q^\T \right) \right\|_{\text F}  \nonumber\\
&= \left\| \mLambda^2 \right\|_{\text F}
\end{align}
Substituting \cref{eq:attnM-gamma} into \cref{eq:attnM-duration}, we obtain the approximate duration of the loss plateau
\begin{align}
t_{\text{plateau}} \approx \frac{\tau}{2\left\| \mLambda^2 \right\|_{\text F}} \ln \frac1{s(0)} 
\approx \frac{\tau}{\left\| \mLambda^2 \right\|_{\text F}} \ln \frac1{w_\init}  ,
\end{align}
where we used the definition $s(0) = \| \vw_2(0) \|^2 = w_\init^2$.

\begin{figure}
  \centering
  \includegraphics[width=\linewidth]{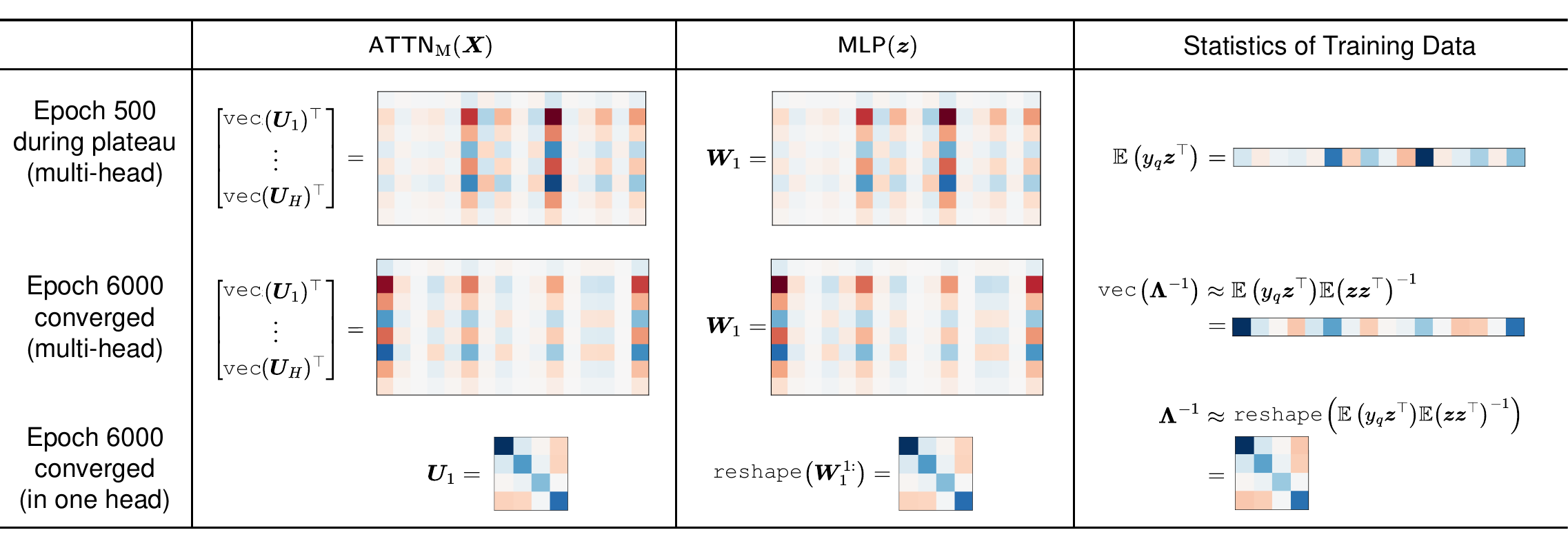}
  \caption{The dynamics of weights in multi-head linear attention with merged key and query can be predicted with statistics of the training dataset. We plot the weights at different times in training, corresponding to the loss trajectories in \cref{fig:attnM} (right).
  The weights in linear attention (first column) stay close to the weights in the fully-connected linear network (second column) throughout training. 
  During the initial plateau, the vectorized key-query weights in attention $\VEC(\mU_1),\cdots,\VEC(\mU_i)$ and the first-layer weight in the fully-connected network $\mW_1$ are rank-one and align with the correlation between the output and the cubic feature input $\E \left(y_q \vz^\T \right)$ (top row). 
  At convergence, $\VEC(\mU_1),\cdots,\VEC(\mU_i)$ in attention and $\mW_1$ in the fully-connected linear network are rank-one and align with the linear regression solution in the cubic feature space $\E \left( y_q \vz^\T \right) \E \left( \vz \vz^\T \right)^{-1}$ (middle row), which is also the in-context linear regression solution in the original token space $\mLambda^{-1}$ (bottom row) as described by \cref{eq:attnM-converged}. The approximate equality in the third column is exact when the sequence length $N\to \infty$.
  }
  \label{fig:attnM-W}
\end{figure}
\subsubsection{Weights Dynamics}
For a white input covariance, the training dynamics reduces to a scalar ordinary differential equation about $s(t)$ given in \cref{eq:ode-s}.
For a general input covariance, the vector $\vm$ in the change of variable defined in \cref{eq:change-variable} rotates during training.
As shown in the top row of \cref{fig:attnM-W}, during the initial loss plateau, the rows of the first-layer weight align with the input-output correlation $\E \left( y_q \vz^\T \right)$ but do not change appreciably in scale \citep{atanasov22silent}. 
Later, when the loss decreases rapidly, the first-layer weight grows in scale and rotates to align with the global minimum solution, $\E \left( y_q \vz^\T \right) \E \left( \vz \vz^\T \right)^{-1}$. 
The alignment and rotation behaviors apply to the rows of the first-layer weight in the fully-connected network, corresponding to the merged key-query weights in the different heads in linear attention, as shown in \cref{fig:attnM-W}.

\subsection{Conservation Law: All Heads Are Parallel}
The weights in a fully-connected linear network are known to obey a conservation law during training \citep{fukumizu98batch,saxe14exact,du18balance,ji18align}
\begin{align}  \label{eq:attnM-conservation}
\frac{\diff}{\diff t} \left( \vw_2 \vw_2^\T - \mW_1 \mW_1^\T \right) = \vzero ,
\end{align} 
which follows directly from the gradient flow dynamics in \cref{eq:attnM-gd}.
Under small initialization, the quantity $\vw_2 \vw_2^\T - \mW_1 \mW_1^\T \approx \vzero$ is small at initialization and remains small throughout training. Since the vector $\vw_2$ is rank-one, the conservation law forces $\mW_1$ to also be approximately rank-one, which means that the rows of $\mW_1$ are approximately parallel. 
Since each row of $\mW_1$ is the vectorized merged key-query matrix of a head, $\VEC(\mU_i)$, a rank-one $\mW_1$ implies that the key-query weight matrices of all heads are parallel, differing only in scale. As shown in \cref{fig:attnM-W}, simulations indeed show that the key-query weights in different heads are parallel.

\section{Linear Attention with Separate Rank-One Key and Query}
\subsection{Justification for Zero Blocks Assumption  \label{supp:attnS-zerouv}}
This is a special case of linear attention with separate rank-$R$ key and query. The proof for the more general rank-$R$ case can be found in  \cref{supp:attnS-lowrank-zerouv}.

\subsection{Gradient Flow Equations  \label{supp:attnS-gd}}
We here derive the gradient flow dynamics for linear attention with separate rank-one key and query introduced in \cref{eq:attnS-gd}.

Based on the gradient flow training rule in \cref{eq:grad-flow}, the gradient flow dynamics for the value, key, and query weights in the $i$-th head are
\begin{subequations}   \label{eq:attnS-gd-raw}
\begin{align}
\tau \dot v_i &=  \vk_i^\T \E \left( \vbeta (y_q-\hat y_q) \vx_q^\T \right) \vq_i  ,\\
\tau \dot \vk_i &= v_i \E \left( \vbeta (y_q-\hat y_q) \vx_q^\T \right) \vq_i  ,\\
\tau \dot \vq_i &= v_i \E \left( \vx_q (y_q-\hat y_q) \vbeta^\T \right) \vk_i  .
\end{align}
\end{subequations}
We calculate the common term in \cref{eq:attnS-gd-raw}, that is
\begin{align}  \label{eq:attnS-grad-Eterm}
\E \left( \vbeta (y_q-\hat y_q) \vx_q^\T \right)
&= \E \left[ \vbeta \left(\vw^\T \vx_q - \sum_{i=1}^H v_i \vbeta^\T \vk_i \vq_i^\T \vx_q \right) \vx_q^\T \right]  \nonumber\\
&= \E \left[ \hat\mLambda \vw \vw^\T \left( \mI - \sum_{i=1}^H v_i \hat\mLambda \vk_i \vq_i^\T \right) \vx_q \vx_q^\T \right]  \nonumber\\
&= \E\left(\hat\mLambda\right) \E_\vw \left(\vw \vw^\T\right) \E_{\vx_q} \left(\vx_q \vx_q^\T\right) 
- \E\left(\hat\mLambda \vw \vw^\T \hat\mLambda\right) \sum_{i=1}^H v_i \vk_i \vq_i^\T  \E_{\vx_q} \left(\vx_q \vx_q^\T\right)   \nonumber\\
&= \mLambda^2 - \E\left({\hat\mLambda}^2\right) \sum_{i=1}^H v_i \vk_i \vq_i^\T \mLambda
\end{align}
Substituting \cref{eq:attnS-grad-Eterm} into \cref{eq:attnS-gd-raw}, we arrive at the same equations as \cref{eq:attnS-gd} in the main text
\begin{subequations}  
\begin{align*}
\tau \dot v_i &=  \vk_i^\T \left( \mLambda^2 - \E\left({\hat\mLambda}^2\right) \sum_{i'=1}^H v_{i'} \vk_{i'} \vq_{i'}^\T \mLambda \right) \vq_i  ,\\
\tau \dot \vk_i &= v_i \left( \mLambda^2 - \E\left({\hat\mLambda}^2\right) \sum_{i'=1}^H v_{i'} \vk_{i'} \vq_{i'}^\T \mLambda \right) \vq_i  ,\\
\tau \dot \vq_i &= v_i \left( \mLambda^2 - \mLambda \sum_{i'=1}^H v_{i'} \vk_{i'} \vq_{i'}^\T \E\left({\hat\mLambda}^2\right) \right) \vk_i  .
\end{align*}
\end{subequations}
where the data statistics $\E\left({\hat\mLambda}^2\right)$ is calculated in \cref{eq:ELambdaLambda}.

\subsection{Fixed Points  \label{supp:attnS-landscape}}
We prove that the fixed points given in \cref{eq:attnS-def-M} are valid.

\begin{proof}
When the model is at a fixed point in set $\gM(\gS_m)$, it satisfies \cref{eq:attnS-condition1}. \cref{eq:attnS-condition1} can be rewritten using $a_d$ (defined in \cref{eq:def-ad}) as
\begin{align}  \label{eq:attnS-at-Mm}
\sum_{i=1}^H v_i \vk_i \vq_i^\T = \sum_{d\in \gS_m} \frac{\lambda_d}{a_d} \ve_d \ve_d^\T .
\end{align}
Using \cref{eq:LambdaLambda-eig,eq:attnS-at-Mm}, we can simplify a common term in the gradient descent dynamics in \cref{eq:attnS-gd} to
\begin{align}  \label{eq:Eterm-at-Mm}
\mLambda^2 - \E\left({\hat\mLambda}^2\right) \sum_{i=1}^H v_i \vk_i \vq_i^\T \mLambda
&= \sum_{d=1}^D \lambda_d^2 \ve_d \ve_d^\T - \sum_{d'=1}^D a_{d'} \ve_{d'} \ve_{d'}^\T \sum_{d \in \gS_m} \frac{\lambda_d}{a_d} \ve_d \ve_d^\T \mLambda  \nonumber\\
&= \sum_{d=1}^D \lambda_d^2 \ve_d \ve_d^\T - \sum_{d \in \gS_m} \lambda_d \ve_d \ve_d^\T \mLambda  \nonumber\\
&= \sum_{d \notin \gS_m} \lambda_d^2 \ve_d \ve_d^\T  .
\end{align}
Substituting \cref{eq:Eterm-at-Mm} into \cref{eq:attnS-gd}, we obtain the dynamics when the model is at a fixed point in $\gM(\gS_m)$
\begin{subequations}  \label{eq:attnS-gd-at-Mm}
\begin{align}
\tau \dot v_i &= \vk_i^\T \left( \sum_{d \notin \gS_m} \lambda_d^2 \ve_d \ve_d^\T \right) \vq_i  , \label{eq:attnS-gd-at-Mm-v} \\
\tau \dot \vk_i &= v_i \left( \sum_{d \notin \gS_m} \lambda_d^2 \ve_d \ve_d^\T \right) \vq_i  , \label{eq:attnS-gd-at-Mm-k} \\
\tau \dot \vq_i &= v_i \left( \sum_{d \notin \gS_m} \lambda_d^2 \ve_d \ve_d^\T \right) \vk_i  . \label{eq:attnS-gd-at-Mm-q}
\end{align}
\end{subequations}
\begin{enumerate}[label=(\roman*)]
    \item For the heads with a nonzero value weight, $v_i\neq 0$, the key and query weights at a fixed point satisfy condition (C2) for \cref{eq:attnS-def-M}, that is the key and query weights lie in the span of $\{ \ve_d \}_{d\in \gS_m}$ and thus can be written as
    \begin{subequations}  \label{eq:kq-in-span}
    \begin{align}
        \vk_i &= \sum_{d\in \gS_m} b_d \ve_d ,\quad b_d \in \sR , \\
        \vq_i &= \sum_{d\in \gS_m} c_d \ve_d ,\quad c_d \in \sR .  \label{eq:q-in-span}
    \end{align}
    \end{subequations}
    Substituting \cref{eq:kq-in-span} into the gradient flow dynamics given in \cref{eq:attnS-gd-at-Mm}, we obtain
    \begin{align*}
    \tau \dot v_i &= \vk_i^\T \left( \sum_{d \notin \gS_m} \lambda_d^2 \ve_d \ve_d^\T \right) \sum_{d'\in \gS_m} c_{d'} \ve_{d'} = 0  ,\\
    \tau \dot \vk_i &= v_i \left( \sum_{d \notin \gS_m} \lambda_d^2 \ve_d \ve_d^\T \right) \sum_{d'\in \gS_m} c_{d'} \ve_{d'} = \vzero  ,\\
    \tau \dot \vq_i &= v_i \left( \sum_{d \notin \gS_m} \lambda_d^2 \ve_d \ve_d^\T \right) \sum_{d'\in \gS_m} b_{d'} \ve_{d'} = \vzero  ,
    \end{align*}
    where we have used the fact that $\ve_d^\T \ve_{d'} =0$ if $d\neq d'$, because eigenvectors of the covariance matrix $\mLambda$ are orthogonal.   
    \item For the heads with a zero value weight, $v_i=0$, the gradients of the key and query weights in \cref{eq:attnS-gd-at-Mm-k,eq:attnS-gd-at-Mm-q} contain $v_i$ and are thus zero, $\dot \vk_i=\vzero, \dot \vq_i=\vzero$. Further, the key and query weights of a head with a zero value weight satisfy condition (C3) for \cref{eq:attnS-def-M}. Without loss of generality, suppose that $\vq_i$ lies in the span of $\{ \ve_d \}_{d\in \gS_m}$, that is $\vq_i$ satisfies \cref{eq:q-in-span}. Substituting \cref{eq:q-in-span} into the gradient of $v_i$ given in \cref{eq:attnS-gd-at-Mm-v}, we obtain
    \begin{align*}
    \dot v_i = \vk_i^\T \left( \sum_{d \notin \gS_m} \lambda_d^2 \ve_d \ve_d^\T \right) \sum_{d'\in \gS_m} c_{d'} \ve_{d'} = 0 ,
    \end{align*}
    where we have again used the fact that eigenvectors of $\mLambda$ are orthogonal.   
\end{enumerate}
Hence, when the model has weights specified in \cref{eq:attnS-def-M}, the gradients of the weights are zero, meaning that the fixed points are valid.
\end{proof}

\subsection{Loss Value at A Fixed Point  \label{supp:attnS-loss}}
We derive the loss when the model is at a fixed point in set $\gM(\gS_m)$, where the loss is given by
\begin{align}  \label{eq:attnS-loss-MSm}
\Ls(\gM(\gS_m)) = \tr(\mLambda) - \sum_{d \in \gS_m} \lambda_d \left( 1 + \frac{1+\tr(\mLambda)/\lambda_d}N \right)^{-1}  .
\end{align}
\cref{eq:attnS-loss} in the main text follows directly from \cref{eq:attnS-loss-MSm} when taking $S_m = \{ 1,2,\cdots, m \}$.

\begin{proof}
We substitute \cref{eq:LambdaLambda-eig,eq:attnS-at-Mm} into the mean square loss and obtain
\begin{align}  \label{eq:attnS-loss-temp}
\Ls(\gM(\gS_m)) = \E (y_q - \hat y_q)^2 
&= \E \left( \vw^\T \vx_q - \sum_{d \in \gS_m} \frac{\lambda_d}{a_d} \vw^\T \hat \mLambda \ve_d \ve_d^\T \vx_q \right)^2  \nonumber\\
&= \E \left[ \vx_q^\T \left( \mI - \sum_{d \in \gS_m} \frac{\lambda_d}{a_d} \hat \mLambda \ve_d \ve_d^\T \right) \E_\vw(\vw \vw^\T) \left( \mI - \sum_{d \in \gS_m} \frac{\lambda_d}{a_d} \hat \mLambda \ve_d \ve_d^\T \right) \vx_q \right]  \nonumber\\
&= \E \left[ \vx_q^\T \left( \mI - \sum_{d \in \gS_m} \frac{\lambda_d}{a_d} \hat \mLambda \ve_d \ve_d^\T \right) \left( \mI - \sum_{d \in \gS_m} \frac{\lambda_d}{a_d} \hat \mLambda \ve_d \ve_d^\T \right) \vx_q \right]  \nonumber\\
&= \E \left[ \vx_q^\T \left( \mI 
-2 \mathcolor{violet}{ \sum_{d \in \gS_m} \frac{\lambda_d}{a_d} \hat \mLambda \ve_d \ve_d^\T }
+\mathcolor{teal}{ \left( \sum_{d \in \gS_m} \frac{\lambda_d}{a_d} \hat \mLambda \ve_d \ve_d^\T \right)^2 }
\right) \vx_q \right] .
\end{align}
Since $\hat\mLambda$ is independent of $\vx_q$, we can calculate the expectation of the {\color{violet}purple} and {\color{teal}teal} terms first,
\begin{align*}
\E \left( \mathcolor{violet}{ \sum_{d \in \gS_m} \frac{\lambda_d}{a_d} \hat \mLambda \ve_d \ve_d^\T } \right)
&= \sum_{d \in \gS_m} \frac{\lambda_d}{a_d} \mLambda \ve_d \ve_d^\T
= \sum_{d \in \gS_m} \frac{\lambda_d^2}{a_d} \ve_d \ve_d^\T ,
\\
\E \left[ \mathcolor{teal}{ \left( \sum_{d \in \gS_m} \frac{\lambda_d}{a_d} \hat \mLambda \ve_d \ve_d^\T \right)^2 } \right]
&= \E \left[ \sum_{d \in \gS_m} \frac{\lambda_d^2}{a_d^2} \ve_d \ve_d^\T \hat \mLambda \hat \mLambda \ve_d \ve_d^\T 
+ \sum_{d,d' \in \gS_m, d \neq d'} \frac{\lambda_d\lambda_{d'}}{a_da_{d'}} \hat \mLambda \ve_d \ve_d^\T \ve_{d'} \ve_{d'}^\T \hat \mLambda  \right]  \\
&= \sum_{d \in \gS_m} \frac{\lambda_d^2}{a_d^2} \ve_d \ve_d^\T \E\left(\hat\mLambda \hat\mLambda \right) \ve_d \ve_d^\T + \vzero  \\
&= \sum_{d \in \gS_m} \frac{\lambda_d^2}{a_d^2} \ve_d \ve_d^\T \sum_{d'=1}^D a_{d'} \ve_{d'} \ve_{d'}^\T \ve_d \ve_d^\T  \\
&= \sum_{d \in \gS_m} \frac{\lambda_d^2}{a_d} \ve_d \ve_d^\T .
\end{align*}
Substituting them back into \cref{eq:attnS-loss-temp}, we get
\begin{align*}
\Ls(\gM(\gS_m))
&= \E \left[ \vx_q^\T \left( \mI - 2 \sum_{d \in \gS_m} \frac{\lambda_d^2}{a_d} \ve_d \ve_d^\T + \sum_{d \in \gS_m} \frac{\lambda_d^2}{a_d} \ve_d \ve_d^\T \right) \vx_q \right]  \\
&= \E \left[ \vx_q^\T \left( \mI - \sum_{d \in \gS_m} \frac{\lambda_d^2}{a_d} \ve_d \ve_d^\T \right) \vx_q \right]  \\
&= \E \left( \vx_q^\T \vx_q \right) - \sum_{d \in \gS_m} \frac{\lambda_d^2}{a_d} \E \left( \vx_q^\T \ve_d \ve_d^\T \vx_q \right)  \\
&= \tr(\mLambda) - \sum_{d \in \gS_m} \frac{\lambda_d^2}{a_d} \ve_d^\T \mLambda \ve_d  \\
&= \tr(\mLambda) - \sum_{d \in \gS_m} \frac{\lambda_d^3}{a_d}
\end{align*}
We plug in the definition of $a_d$ in \cref{eq:def-ad} and arrive at the desired result:
\begin{align*}
\Ls(\gM(\gS_m)) &= \tr(\mLambda) - \sum_{d \in \gS_m} \lambda_d^3 \frac1{\lambda_d^2} \left( 1 + \frac{1+\tr(\mLambda)/\lambda_d}N \right)^{-1}  \\
&= \tr(\mLambda) - \sum_{d \in \gS_m} \lambda_d \left( 1 + \frac{1+\tr(\mLambda)/\lambda_d}N \right)^{-1}  .
\end{align*}
\end{proof}

\subsection{Saddle-to-Saddle Dynamics: From $\gM_0$ to $\gM_1$  \label{supp:M0-M1}}
\begin{figure}[h]
  \centering
  \includegraphics[width=0.39\linewidth]{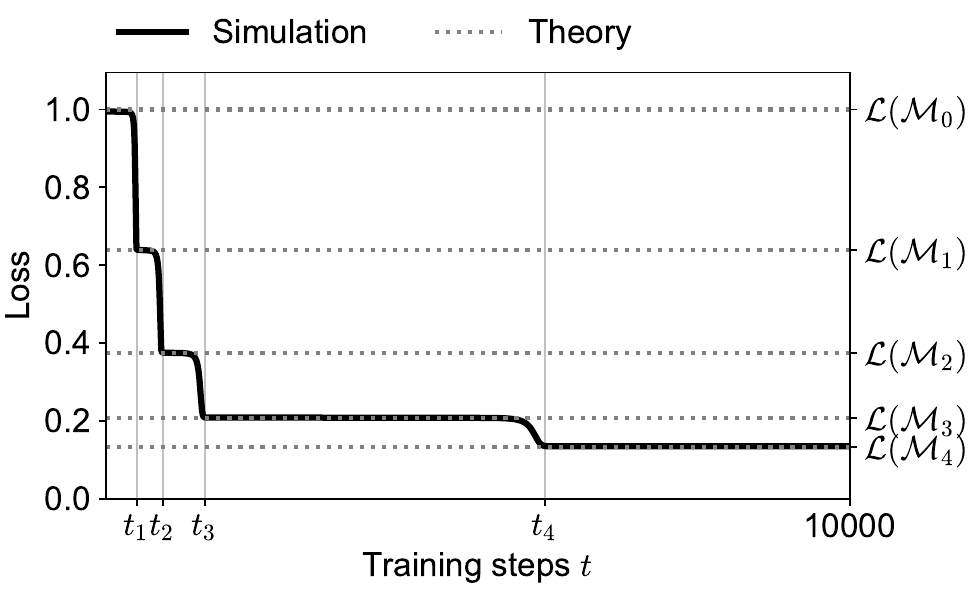}
  \caption{Illustration of $t_1,\cdots,t_D$. The loss trajectory plotted is one of the trajectories of linear attention with separate rank-one key and query in \cref{fig:attnS-loss}. The time $t_d \,(d=1,\dots,D)$ denotes the time when the loss has just undergone the $d$-th abrupt drop.}
  \label{fig:attnS-timedemo}
\end{figure}

We denote the time at which the loss has just undergone the $d$-th abrupt drop as $t_d \,(d=1,\dots,D)$, as illustrated in \cref{fig:attnS-timedemo}.

\subsubsection{Alignment During the Plateau.}
In the initial loss plateau, the weights have not moved much away from their small initialization and thus the training dynamics are mainly driven by the first terms in \cref{eq:attnS-gd}, which are
\begin{subequations}  \label{eq:attnS-gd-early}
\begin{align}
\tau \dot v_i &=  \vk_i^\T \mLambda^2 \vq_i + O(w_\init^5) ,  \label{eq:attnS-gd-v-early} \\
\tau \dot \vk_i &= v_i \mLambda^2 \vq_i + O(w_\init^5) ,  \\
\tau \dot \vq_i &= v_i \mLambda^2 \vk_i + O(w_\init^5) .
\end{align}
\end{subequations}
With a small initialization scale $w_\init$, the key and query weights in a head evolve approximately as
\begin{align}  \label{eq:attnS-gd-kq-early}
\tau \frac{\diff }{\diff t} \begin{bmatrix}
\vk_i  \\  \vq_i
\end{bmatrix}
= v_i \begin{bmatrix}
\vzero  &  \mLambda^2  \\
\mLambda^2  &  \vzero
\end{bmatrix}  \begin{bmatrix}
\vk_i  \\  \vq_i
\end{bmatrix}  .
\end{align}
The matrix $\begin{bmatrix}
\vzero  &  \mLambda^2  \\
\mLambda^2  &  \vzero
\end{bmatrix} \in \sR^{2D \times 2D}$ has eigenvalues $\left\{ \lambda_d^2, -\lambda_d^2 \right\}_{d=1}^D$, corresponding to eigenvectors
\begin{align*}
\begin{bmatrix}
\vzero  &  \mLambda^2  \\
\mLambda^2  &  \vzero
\end{bmatrix} \begin{bmatrix}
\ve_d \\ \ve_d
\end{bmatrix}
=  \lambda_d^2 \begin{bmatrix}
\ve_d \\ \ve_d
\end{bmatrix}
,\quad
\begin{bmatrix}
\vzero  &  \mLambda^2  \\
\mLambda^2  &  \vzero
\end{bmatrix} \begin{bmatrix}
\ve_d \\ -\ve_d
\end{bmatrix}
=  -\lambda_d^2 \begin{bmatrix}
\ve_d \\ -\ve_d
\end{bmatrix}
,\quad
d=1,\cdots,D.
\end{align*}
where recall that $\lambda_d,\ve_d (d=1,\cdots,D)$ are eigenvalues and eigenvectors of $\mLambda$. 
Hence, the solution to \cref{eq:attnS-gd-kq-early} takes the following form
\begin{align}  \label{eq:attnS-early-sol}
\begin{split}
\begin{bmatrix}
\vk_i(t)  \\  \vq_i(t)
\end{bmatrix}
&= \frac12 \sum_{d=1}^D \ve_d^\T\left(\vk_i(0)+\vq_i(0)\right) \exp\left( \frac{\lambda_d^2}{\tau} \int_0^t v_i(t') \diff t' \right) \begin{bmatrix}
\ve_d \\ \ve_d
\end{bmatrix}  \\
&+ \frac12 \sum_{d=1}^D \ve_d^\T \left(\vk_i(0)-\vq_i(0)\right) \exp\left( -\frac{\lambda_d^2}{\tau} \int_0^t v_i(t') \diff t' \right) \begin{bmatrix}
\ve_d \\ -\ve_d
\end{bmatrix}  .
\end{split}
\end{align}
If $v_i>0$, the first summation term in \cref{eq:attnS-early-sol} grows and the second summation term decays. The key and query weights $\vk_i,\vq_i$ both grow in size along the directions of the eigenvectors $\ve_d$.
If $v_i<0$, the first summation term in \cref{eq:attnS-early-sol} decays and the second summation term grows. The key and query weights $\vk_i,\vq_i$ grow in opposite directions, $\ve_d$ and $-\ve_d$ respectively. 
In either case, the multiplication $v_i \vk_i \vq_i^\T$ grows along $\ve_d \ve_d^\T$. 

\subsubsection{Reduction to Scalar Dynamics with An Alignment Ansatz.  \label{supp:attnS-duration}}
The dominating term in \cref{eq:attnS-early-sol} is the term with the largest positive eigenvalue. In other words, the key and query weights grow the fastest along the first eigenvector $\ve_1$ and thus are approximately aligned with $\ve_1$.
Motivated by this insight, we make an ansatz that the key and query weights in a head are exactly aligned with $\ve_1$ and the rest of the heads are zero\footnote{We let the head aligned with $\ve_1$ to have index 1.}:
\begin{subequations}  \label{eq:attnS-ansatz}
\begin{align}
\vk_1 &= \vq_1 = v_1 \ve_1  ,  \\
\vk_i &= \vq_i = \vzero, v_i = 0 ,\, i = 2,\cdots, H.
\end{align}
\end{subequations}
Note that \cref{eq:attnS-ansatz} also assumes that the $\ell^2$ norms of $\vk_1, \vq_1, v_1$ are equal, which is true under vanishing initialization due to the conservation law in \cref{eq:attnS-conservation}. This ansatz can greatly simplify the training dynamics and provide a good approximation of the true dynamics, where weights in one of the heads grow in scale with the key and query weights aligning with $\ve_1$, while the rest of the heads remain near zero from time $0$ to $t_1$.

We substitute the ansatz into the training dynamics in \cref{eq:attnS-gd} to reduce the high-dimensional dynamics to a one-dimensional ordinary differential equation. To do that, we first calculate the common expectation term in the training dynamics with the ansatz,
\begin{align}  \label{eq:Eterm-at-M0}
\mLambda^2 - \E\left({\hat\mLambda}^2\right) \sum_{i=1}^H v_i \vk_i \vq_i^\T \mLambda
= \mLambda^2 - \sum_{d=1}^D a_d \ve_d \ve_d^\T v_1^3 \ve_1 \ve_1^\T \mLambda  
= \mLambda^2 - \lambda_1 a_1 \ve_1 \ve_1^\T v_1^3 
\end{align}
where $a_1$ is the first eigenvalue of $\E({\hat\mLambda}^2)$ defined in \cref{eq:def-ad}.
Substituting \cref{eq:attnS-ansatz,eq:Eterm-at-M0} into \cref{eq:attnS-gd}, we find that the training dynamics of the first head simplify and the dynamics of the rest of the heads are zero
\begin{align*}
\tau \dot v_1 &= v_1^2 \ve_1^\T \left( \mLambda^2 - \lambda_1 a_1 \ve_1 \ve_1^\T v_1^3 \right) \ve_1
= \lambda_1^2 v_1^2 - \lambda_1 a_1 v_1^5  ,  \\
\tau \dot \vk_1 &= v_1^2 \left( \mLambda^2 - \lambda_1 a_1 \ve_1 \ve_1^\T v_1^3  \right) \ve_1
= \lambda_1^2 v_1^2 \ve_1 - \lambda_1 a_1 v_1^5 \ve_1  ,  \\
\tau \dot \vq_1 &= v_1^2 \left( \mLambda^2 - \lambda_1 a_1 \ve_1 \ve_1^\T v_1^3  \right) \ve_1
= \lambda_1^2 v_1^2 \ve_1 - \lambda_1 a_1 v_1^5 \ve_1  , \\
\dot v_i &= 0, \dot \vk_i = \vzero, \dot \vq_i = \vzero ,\, i=2,\cdots,H. 
\end{align*}
We further substitute in $\dot \vk_1 = \dot v_1 \ve_1, \dot \vq_1 = \dot v_1 \ve_1$ and find that the high-dimensional training dynamics reduce to one-dimensional dynamics about $v_1(t)$
\begin{align}  \label{eq:1Ddynamics-v1}
\begin{cases}
\tau \dot v_1 = \lambda_1^2 v_1^2 - \lambda_1 a_1 v_1^5  \\
\tau \dot v_1 \ve_1 = \lambda_1^2 v_1^2 \ve_1 - \lambda_1 a_1 v_1^5 \ve_1  \\
\tau \dot v_1 \ve_1 = \lambda_1^2 v_1^2 \ve_1 - \lambda_1 a_1 v_1^5 \ve_1 
\end{cases}
\quad \Rightarrow \quad
\tau \dot v_1 = \lambda_1^2 v_1^2 - \lambda_1 a_1 v_1^5
\end{align}
\cref{eq:1Ddynamics-v1} is a separable ordinary differential equation. By separating variables and integrating both sides, we can solve $t$ in terms of $v_1$
\begin{align}  \label{eq:wolfram-solution}
\frac{\lambda_1^2}{\tau} t &= \int \frac{1}{v_1^2 - \frac{a_1}{\lambda_1}v_1^2} \diff v_1  \nonumber\\
&= \frac{\sqrt[3]{\frac{a_1}{\lambda_1}}}6 
\left[ \ln \left( \frac{\sqrt[3]{\frac{a_1^2}{\lambda_1^2}}v_1^2+\sqrt[3]{\frac{a_1}{\lambda_1}}v_1+1}
{\sqrt[3]{\frac{a_1^2}{\lambda_1^2}}v_1^2-2\sqrt[3]{\frac{a_1}{\lambda_1}}v_1+1} \right)
- 2\sqrt3 \tan^{-1} \left(\frac{2\sqrt[3]{\frac{a_1}{\lambda_1}}v_1+1}{\sqrt3} \right) \right] 
- \frac1{v_1}  .
\end{align}
Since \cref{eq:wolfram-solution} does not have a straight-forward inverse, we cannot obtain a general analytical solution of $v_1(t)$ in terms of $t$. Nonetheless, we can readily generate numerical solutions and obtain approximate analytical solutions when $v_1$ is near its small initialization to estimate the duration of the first loss plateau.

When $v_1$ is small, the dominating term in \cref{eq:1Ddynamics-v1} is $\lambda_1^2 v_1^2$ and thus the dynamics can be approximated by
\begin{align*}
\tau \dot v_i &= \lambda_1^2 v_i^2
\quad \Rightarrow \quad
t = \frac{\tau}{\lambda_1^2} \left( \frac1{v_i(0)} - \frac1{v_i(t)} \right)  .
\end{align*}
At the end of the plateau, $v_1(t)$ has grown to be much larger than $v_1(0)$. Hence, the duration of the first loss plateau, $t_1$, is
\begin{align}  \label{eq:attnS-t1}
t_1 \approx \frac{\tau}{\lambda_1^2 v_1(0)}  .
\end{align}

\subsection{Saddle-to-Saddle Dynamics: From $\gM_m$ to $\gM_{m+1}$  \label{supp:M1-M2}}
In \cref{supp:M0-M1}, we have analyzed the training dynamics from time $0$ to $t_1$, during which the model moves from saddle $\gM_0$ to saddle $\gM_1$. 
We now analyze the general saddle-to-saddle dynamics from time $t_m$ to $t_{m+1} \,(m=0,\cdots,D-1)$, during which the model moves from $\gM_m$ to $\gM_{m+1}$.

\subsubsection{Alignment During the Plateau.}
Based on our dynamics analysis from time $0$ to $t_1$ and by induction, the weights during the $m$-th plateau are approximately described by \cref{eq:attnS-Mm-ansatz}. Namely, there are $m$ heads whose key and query weights have grown and become aligned with the first $m$ eigenvectors while weights in the rest of the heads have not moved much from their small initialization. Thus, similarly to \cref{eq:attnS-gd-at-Mm}, the heads that are near small initialization have the following training dynamics
\begin{align*}
\tau \dot v_i &= \vk_i^\T \left( \sum_{d=m+1}^D \lambda_d^2 \ve_d \ve_d^\T \right) \vq_i + O(w_\init^5)  , \\
\tau \dot \vk_i &= v_i \left( \sum_{d=m+1}^D \lambda_d^2 \ve_d \ve_d^\T \right) \vq_i + O(w_\init^5)  , \\
\tau \dot \vq_i &= v_i \left( \sum_{d=m+1}^D \lambda_d^2 \ve_d \ve_d^\T \right) \vk_i + O(w_\init^5)  . 
\end{align*}
With a small initialization scale $w_\init$, the key and query weights in this head evolve approximately as
\begin{align}  \label{eq:attnS-gd-kq-plateau}
\tau \frac{\diff }{\diff t} \begin{bmatrix}
\vk_i  \\  \vq_i
\end{bmatrix}
= v_i \begin{bmatrix}
\vzero & \mOmega   \\
\mOmega & \vzero
\end{bmatrix}  \begin{bmatrix}
\vk_i  \\  \vq_i
\end{bmatrix} 
,\quad \text{where }
\mOmega = \sum_{d=m+1}^D \lambda_d^2 \ve_d \ve_d^\T .
\end{align}
The matrix $\begin{bmatrix}
\vzero  &  \mOmega  \\
\mOmega  &  \vzero
\end{bmatrix} \in \sR^{2D \times 2D}$ has $2m$ zero eigenvalues and $(2D-2m)$ nonzero eigenvalues, which are $\left\{ \lambda_d^2, -\lambda_d^2 \right\}_{d=m+1}^D$. The nonzero eigenvalues correspond to eigenvectors
\begin{align*}
\begin{bmatrix}
\vzero  &  \mOmega  \\
\mOmega  &  \vzero
\end{bmatrix} \begin{bmatrix}
\ve_d \\ \ve_d
\end{bmatrix}
=  \lambda_d^2 \begin{bmatrix}
\ve_d \\ \ve_d
\end{bmatrix}
,\quad
\begin{bmatrix}
\vzero  &  \mOmega  \\
\mOmega  &  \vzero
\end{bmatrix} \begin{bmatrix}
\ve_d \\ -\ve_d
\end{bmatrix}
=  -\lambda_d^2 \begin{bmatrix}
\ve_d \\ -\ve_d
\end{bmatrix}
,\quad
d=m+1,\cdots,D.
\end{align*}
Hence, the solution to \cref{eq:attnS-gd-kq-plateau} takes the following the form
\begin{align}    \label{eq:attnS-tm-sol}
\begin{split}
\begin{bmatrix}
\vk_i(t)  \\  \vq_i(t)
\end{bmatrix}
&= \frac12 \sum_{d=m+1}^D \ve_d^\T\left(\vk_i(t_m)+\vq_i(t_m)\right) \exp\left( \frac{\lambda_d^2}{\tau} \int_{t_m}^t v_i(t') \diff t' \right) \begin{bmatrix}
\ve_d \\ \ve_d
\end{bmatrix}  \\
&+ \frac12 \sum_{d=m+1}^D \ve_d^\T\left(\vk_i(t_m)-\vq_i(t_m)\right) \exp\left( -\frac{\lambda_d^2}{\tau} \int_{t_m}^t v_i(t') \diff t' \right) \begin{bmatrix}
\ve_d \\ -\ve_d
\end{bmatrix} \\ 
&+ \sum_{d=1}^m \ve_d^\T \left(\vk_i(t_m)+\vq_i(t_m)\right) \begin{bmatrix}
\ve_d \\ \ve_d
\end{bmatrix}  .
\end{split}
\end{align}
For $v_i>0$, the first term grows and the second term decays with time. The third term does not change with respect to time.

\subsubsection{Reduction to Scalar Dynamics with An Alignment Ansatz.}
The dominating term in \cref{eq:attnS-tm-sol} is the term with the largest positive eigenvalue. In other words, during the $(m+1)$-th plateau, the key and query weights that are still near small initialization grow the fastest along the $(m+1)$-th eigenvector $\ve_{m+1}$. Based on this insight, we make the ansatz in \cref{eq:attnS-Mm-ansatz}.
This ansatz can reduce the high-dimensional training dynamics to a one-dimensional ordinary differential equation and provides a good approximation of the true dynamics, where weights in one of the heads grow in scale with the key and query weights aligning with $\ve_{m+1}$, while the rest of the heads do not change much from time $t_m$ to $t_{m+1}$.

To calculate the training dynamics in \cref{eq:attnS-gd} with the ansatz, we first calculate a common term with the ansatz
\begin{align}  \label{eq:Eterm-at-Mm-ansatz}
\mLambda^2 - \E\left({\hat\mLambda}^2\right) \sum_{i=1}^H v_i \vk_i \vq_i^\T \mLambda
&= \mLambda^2 - \sum_{d=1}^D a_d \ve_d \ve_d^\T \left( \sum_{i=1}^m \frac{\lambda_d}{a_d} \ve_i \ve_i^\T + v_{m+1}^3 \ve_{m+1} \ve_{m+1}^\T \right) \mLambda  \nonumber\\
&= \mLambda^2 - \sum_{d=1}^m \lambda_d^2 \ve_d \ve_d^\T -  \lambda_{m+1} a_{m+1} \ve_{m+1} \ve_{m+1}^\T v_{m+1}^3
\end{align}
By substituting \cref{eq:Eterm-at-Mm-ansatz,eq:attnS-Mm-ansatz} into \cref{eq:attnS-gd}, we find that the dynamics for the heads with index $i\neq m+1$ are zero
\begin{align*}
\dot v_i &= 0, \dot \vk_i = \vzero, \dot \vq_i = \vzero ,\, i\neq m+1  . 
\end{align*}
For the head with index $i=m+1$, the dynamics reduce to one-dimensional dynamics about $v_i(t)$
\begin{align}
\tau \dot v_i &= v_i^2 \ve_{m+1}^\T \left( \mLambda^2 - \sum_{d=1}^m \lambda_d^2 \ve_d \ve_d^\T - \lambda_{m+1} a_{m+1} \ve_{m+1} \ve_{m+1}^\T v_i^3 \right) \ve_{m+1}  \nonumber\\
&= \lambda_{m+1}^2 v_i^2 - \lambda_{m+1} a_{m+1} v_i^5   \nonumber\\
\tau \dot \vk_i = \tau \dot v_i \ve_{m+1}
&= v_i^2 \left( \mLambda^2 - \sum_{d=1}^m \lambda_d^2 \ve_d \ve_d^\T - \lambda_{m+1} a_{m+1} \ve_{m+1} \ve_{m+1}^\T v_i^3 \right) \ve_{m+1}  \nonumber\\
&= \lambda_{m+1}^2 v_i^2 \ve_{m+1} - \lambda_{m+1} a_{m+1} v_i^5 \ve_{m+1}  \nonumber\\
\tau \dot \vq_i = \tau \dot v_i \ve_{m+1}
&= v_i^2 \left( \mLambda^2 - \sum_{d=1}^m \lambda_d^2 \ve_d \ve_d^\T - \lambda_{m+1} a_{m+1} \ve_{m+1} \ve_{m+1}^\T v_i^3 \right) \ve_{m+1}   \nonumber\\
&= \lambda_{m+1}^2 v_i^2 \ve_{m+1} - \lambda_{m+1} a_{m+1} v_i^5 \ve_{m+1}  \nonumber\\ 
\Rightarrow \quad
\tau \dot v_i &= \lambda_{m+1}^2 v_i^2 - \lambda_{m+1} a_{m+1} v_i^5   \label{eq:1Ddynamics-vi}
\end{align}
\cref{eq:1Ddynamics-vi} is the same ordinary differential equation as \cref{eq:1Ddynamics-v1} modulo the constant coefficients. Therefore, with the same analysis, we can estimate the duration of the $(m+1)$-th loss plateau.

When $v_{m+1}$ is small, the dominating term in \cref{eq:1Ddynamics-vi} is $\lambda_{m+1}^2 v_i^2$ and thus the dynamics is well approximated by
\begin{align*}
\tau \dot v_{m+1} &= \lambda_{m+1}^2 v_{m+1}^2
\quad \Rightarrow \quad
t - t_m = \frac{\tau}{\lambda_{m+1}^2} \left( \frac1{v_{m+1}(t_m)} - \frac1{v_{m+1}(t)} \right)  .
\end{align*}
At the end of the plateau, $v_{m+1}(t_{m+1})$ has grown to be much larger than $v_{m+1}(t_m)$. Hence, the duration of the $(m+1)$-th loss plateau is
\begin{align}  \label{eq:attnS-tm}
t_{m+1} - t_m \approx \frac{\tau}{\lambda_{m+1}^2 v_{m+1}(t_m)}  .
\end{align}
We note that the \cref{eq:attnS-tm} involves $v_{m+1}(t_m)$, which depends on the random initialization and the dynamics from time $0$ to $t_m$. This explains why we observe the variance of $t_m$ increases with a larger $m$, that is the timing of a later abrupt loss drop varies more across random seeds as shown in \cref{fig:attnS-loss}.

\subsection{Weight Configuration with Minimal L2 Norm}
We prove that \cref{eq:attnS-Mm-ansatz} with $v_{m+1}=0$ is the weight configuration with minimal $\ell^2$ norm that satisfies \cref{eq:attnS-condition1}. To do this, we find the weight configuration with minimal $\ell^2$ norm satisfying a general equality constrain and apply the solution to \cref{eq:attnS-condition1}.

Consider the equality constrained optimization problem
\begin{align*}
\text{minimize } &\sum_{i=1}^H v_i^2 + \| \vk_i\|^2 + \| \vq_i\|^2  \\
\text{subject to } &\sum_{i=1}^H v_i \vk_i \vq_i^\T = \mA
\end{align*}
where $\mA$ is a positive semi-definite matrix.

\begin{proof}
We use Lagrange multiplier to solve this equality constrained optimization problem.
First, we construct the Lagrangian function $L(\mM)$ where the Lagrange multiplier $\mM \in \sR^{D\times D}$ is a symmetric matrix
\begin{align*}
L(\mM) &= \frac12 \sum_{i=1}^H \left( v_i^2 + \| \vk_i\|^2 + \| \vq_i\|^2 \right) + \VEC(\mM)^\T \VEC \left( \mA - \sum_{i=1}^H v_i \vk_i \vq_i^\T \right)  \\
&= \frac12 \sum_{i=1}^H \left( v_i^2 + \| \vk_i\|^2 + \| \vq_i\|^2 \right) + \tr \left[ \mM \left( \mA - \sum_{i=1}^H v_i \vk_i \vq_i^\T \right) \right]
\end{align*}
Differentiating the Lagrangian with respect to all the variables and setting them to zero, we get
\begin{subequations}  \label{eq:lagrange}
\begin{align}
\frac{\partial L}{\partial v_i} &= 
v_i - \vk_i^\T \mM \vq_i
=0  \label{eq:lagrange-vi} \\
\frac{\partial L}{\partial \vk_i} &=
\vk_i - v_i \mM \vq_i
= \vzero  \label{eq:lagrange-ki} \\
\frac{\partial L}{\partial \vq_i} &=
\vq_i - v_i \mM \vk_i
= \vzero  \label{eq:lagrange-qi} \\
\frac{\partial L}{\partial \mM} &= \mA - \sum_{i=1}^H v_i \vk_i \vq_i^\T
=0  \label{eq:lagrange-xi}
\end{align}
\end{subequations}
\cref{eq:lagrange} suggests that, for each head, the value, key, and query weights are either all zero or satisfy a constraint; i.e., for each $i$, either $v_i = \vk_i = \vq_i = 0$ or
\begin{align}   \label{eq:Lagrange-eighead-raw}
\vk_i = v_i \mM \vq_i = v_i^2 \mM^2 \vk_i .
\end{align}
We got \cref{eq:Lagrange-eighead-raw} by substituting \cref{eq:lagrange-qi} into \cref{eq:lagrange-ki}. \cref{eq:Lagrange-eighead-raw} implies that $\vk_i$ is an eigenvector of $\mM^2$. Let us denote the normalized eigenvector of $\mM^2$ as $\bm{\xi}_i$. Substituting \cref{eq:Lagrange-eighead-raw} into \cref{eq:lagrange-vi} and rearranging, we get
\begin{align} \label{eq:Lagrange-eighead}
\vk_i = \vq_i = v_i \bm{\xi}_i .
\end{align}
With \cref{eq:Lagrange-eighead,eq:lagrange-xi}, we obtain
\begin{align}
\mA = \sum_i v_i^3 \bm{\xi}_i \bm{\xi}_i^\T
\quad \Rightarrow \quad
v_i= \lambda_i^{1/3} ,\,  \bm{\xi}_i = \ve_i ,
\end{align}
where $\lambda_i,\ve_i$ are the eigenvalue and eigenvector of $\mA$.

For the optimization problem, the solution is that there are $\texttt{rank}(\mA)$ heads with nonzero weights and $(H-\texttt{rank}(\mA))$ heads with zero weights. The nonzero heads have weights
\begin{align}  \label{eq:lagrange-solution}
\vk_i = \vq_i = v_i \ve_i ,\, v_i = \lambda_i^{1/3}
,\quad i = 1,\cdots, \texttt{rank}(\mA) .
\end{align}
The indices of heads can be trivially permuted. The signs of any two among $v_i,\vk_i,\vq_i$ can be flipped without affecting the optimization problem.
\end{proof}
We apply the solution in \cref{eq:lagrange-solution} to find a weight configuration with the minimal $\ell^2$ norm that satisfies \cref{eq:attnS-condition1}. \cref{eq:attnS-condition1} can be rewritten as \cref{eq:attnS-at-Mm}, namely
\begin{align*}
\sum_{i=1}^H v_i \vk_i \vq_i^\T = \sum_{d\in \gS_m} \frac{\lambda_d}{a_d} \ve_d \ve_d^\T .
\end{align*}
The matrix on the right hand side has rank $m$ and eigenvectors $\ve_d$ with eigenvalues $\lambda_d/a_d$ $(d\in \gS_m)$. Hence, the weight configuration with minimal $\ell^2$ norm has $(H-m)$ heads with zero weights and $m$ heads with nonzero weights. The nonzero heads have weights
\begin{align*}
\vk_i = \vq_i = v_i \ve_i ,\, v_i = \left( \frac{\lambda_d}{a_d} \right)^{\frac13} = \lambda_i^{-\frac13} \left( 1 + \frac{1+\tr(\mLambda)/\lambda_i}N \right)^{-\frac13}
,\quad i = 1,\cdots, m .
\end{align*}
This is the same weight configuration as \cref{eq:attnS-Mm-ansatz} with $v_{m+1}=0$.

\subsection{Conservation Law  \label{supp:attnS-conservation}}
The gradient flow dynamics of linear attention with separate rank-one key and query in \cref{eq:attnS-gd} implies a conservation law.
The value, key, and query weights in a head obey
\begin{align}  \label{eq:attnS-conservation}
\frac{\diff}{\diff t} \left( \vk_i^\T \vk_i - \vq_i^\T \vq_i \right) = \vzero ,\quad
\frac{\diff}{\diff t} \left( \vk_i^\T \vk_i - v_i^2 \right) = 0 ,
\end{align}
Under small initialization, the quantities $\vk_i^\T \vk_i - \vq_i^\T \vq_i \approx \vzero$ and $\vk_i^\T \vk_i - v_i^2 \approx 0$ are small at initialization and remain small throughout training. Thus, the conservation law enforces the $\ell^2$ norms of the value, key, and query to be approximately the same throughout training, $\|\vk_i\|^2\approx \|\vq_i\|^2\approx v_i^2$. 

We here prove that \cref{eq:attnS-conservation} holds regardless of the choice of the loss function.
\begin{proof}
We can use the generic gradient flow equation in \cref{eq:grad-flow} to calculate the gradients of $\vk_i^\T \vk_i,\vq_i^\T \vq_i$, and $v_i^2$,
\begin{align*}
\frac{\diff \vk_i^\T \vk_i}{\diff t} &= 2 \vk_i^\T \frac{\diff \vk_i}{\diff t} 
= 2 \E \left( - \vk_i^\T \frac{\diff \Ls}{\diff \hat y_q} \frac{\diff \hat y_q}{\diff \vk_i} \right)
= 2 \E \left( - \frac{\diff \Ls}{\diff \hat y_q} v_i \vk_i^\T \vbeta \vq_i^\T \vx_q  \right)
\\
\frac{\diff \vq_i^\T \vq_i}{\diff t} &= 2\vq_i^\T \frac{\diff \vq_i}{\diff t} 
= 2 \E \left( - \vq_i^\T \frac{\diff \Ls}{\diff \hat y_q} \frac{\diff \hat y_q}{\diff \vq_i} \right) 
= 2 \E \left( - \frac{\diff \Ls}{\diff \hat y_q} v_i \vq_i^\T \vx_q \vk_i^\T \vbeta \right) 
\\
\frac{\diff v_i^2}{\diff t} &= 2 v_i \frac{\diff v_i}{\diff t}
= 2 \E \left( - v_i \frac{\diff \Ls}{\diff \hat y_q} \frac{\diff \hat y_q}{\diff v_i} \right) 
= 2 \E \left( - \frac{\diff \Ls}{\diff \hat y_q} v_i \vbeta^\T \vk_i \vq_i^\T \vx_q \right) 
\end{align*}
We see that the gradients of $\vk_i^\T \vk_i,\vq_i^\T \vq_i$, and $v_i^2$ are equal, regardless of the specific choice of the loss function $\Ls$.
Hence, the following conservation law holds for any loss function:
\begin{align*}
\frac{\diff}{\diff t} \left( \vk_i^\T \vk_i - \vq_i^\T \vq_i \right) = \vzero ,\quad
\frac{\diff}{\diff t} \left( \vk_i^\T \vk_i - v_i^2 \right) = 0 .
\end{align*}
\end{proof}

\section{Linear Attention with Separate Low-Rank Key and Query}
\subsection{Justification for Zero Blocks Assumption   \label{supp:attnS-lowrank-zerouv}}
We initialize $\vv_i=\vzero,k_{i,r}=0 \, (i=1,\cdots,H,r=1,\cdots,R)$, and prove that they will stay zero throughout training.

\begin{proof}
The bottom right entry of the output of linear attention with separate rank-$R$ key and query is
\begin{align*}
\hat y_q &\equiv \attn_{\text S}(\mX)_{D+1,N+1}  \\
&= \sum_{i=1}^H \begin{bmatrix}
\vv_i^\T & v_i
\end{bmatrix}
\begin{bmatrix}
\frac1N \left(\vx_q \vx_q^\T + \sum_n \vx_n \vx_n^\T \right) & \frac1N \sum_n \vx_n y_n  \\
\frac1N \sum_n y_n \vx^\T & \frac1N \sum_n y_n^2
\end{bmatrix}
\begin{bmatrix}
\vk_{i,1} & \cdots & \vk_{i,R}  \\
k_{i,1} & \cdots & k_{i,R}
\end{bmatrix}
\begin{bmatrix}
\vq_{i,1}^\T \\ \vdots \\  \vq_{i,R}^\T
\end{bmatrix} \vx_q  \\
&= \sum_{i=1}^H \left( \vv_i^\T \left( \hat \mLambda + \frac1N \vx_q \vx_q^\T \right) \sum_{r=1}^R \vk_{i,r} \vq_{i,r}^\T + v_i \vbeta^\T \sum_{r=1}^R \vk_{i,r} \vq_{i,r}^\T 
+ \vv_i^\T \vbeta \sum_{r=1}^R k_{i,r} \vq_{i,r}^\T + v_i \vw^\T \hat\mLambda \vw \sum_{r=1}^R k_{i,r} \vq_{i,r}^\T \right) \vx_q
\end{align*}
If we initialize $\vv_i=\vzero,k_{i,r}=0$, $\hat y_q$ is
\begin{align*}
\hat y_q = \sum_{i=1}^H \sum_{r=1}^R v_i \vbeta^\T \vk_{i,r} \vq_{i,r}^\T \vx_q 
= \vw^\T \hat\mLambda \sum_{i=1}^H \sum_{r=1}^R v_i \vk_{i,r} \vq_{i,r}^\T \vx_q .
\end{align*}
We now calculate the gradient updates of $\vv_i=\vzero,k_{i,r}=0$ and prove their gradients are zero if their initialization is zero. 
The gradient update of $\vv_i$ contains $\E(\vw)$, which is zero. Similarly to \cref{eq:attnM-zerov}, we have
\begin{align*}
\tau \dot \vv_i &= \E \left[ (y_q - \hat y_q) \left( \left( \hat \mLambda + \frac1N \vx_q \vx_q^\T \right) \sum_{r=1}^R \vk_{i,r} \vq_{i,r}^\T + \vbeta \sum_{r=1}^R k_{i,r} \vq_{i,r}^\T \right) \vx_q \right]  \\
&= \E \left[ \left( \vw^\T \vx_q - \vw^\T \hat\mLambda \sum_{i=1}^H \sum_{r=1}^R v_i \vk_{i,r} \vq_{i,r}^\T \vx_q \right) \left( \hat \mLambda + \frac1N \vx_q \vx_q^\T \right) \sum_{r=1}^R \vk_{i,r} \vq_{i,r}^\T \vx_q \right]  \\
&= \E_\vw (\vw)^\T \E \left[ \left( \vx_q - \hat\mLambda \sum_{i=1}^H \sum_{r=1}^R v_i \vk_{i,r} \vq_{i,r}^\T \vx_q \right) \left( \hat \mLambda + \frac1N \vx_q \vx_q^\T \right) \sum_{r=1}^R \vk_{i,r} \vq_{i,r}^\T \vx_q \right]  \\
&= \vzero  .
\end{align*}
The gradient update of $k_{i,r}$ contains $\E_\vw \left( \vw^\T \hat \mLambda \vw \vw^\T \right)$, whose entries are linear combinations of third moments the zero-mean normal random variable $\vw$, and are thus zero. Similarly to \cref{eq:attnM-zerou}, we have
\begin{align*}
\tau \dot k_{i,r} &= \E \left[ \left( \vv_i^\T \vbeta + v_i \vw^\T \hat\mLambda \vw \right) (y_q - \hat y_q) \vq_{i,r}^\T \vx_q \right]  \\
&= \E \left[ v_i \vw^\T \hat \mLambda \vw \left( \vw^\T \vx_q - \vw^\T \hat\mLambda \sum_{i=1}^H \sum_{r'=1}^R v_i \vk_{i,r'} \vq_{i,r'}^\T \vx_q \right) \vq_{i,r}^\T \vx_q \right]  \\
&= \E_\vw \left( \vw^\T \hat \mLambda \vw \vw^\T \right) \E \left[ v_i \left( \vx_q - \hat\mLambda \sum_{i=1}^H \sum_{r'=1}^R v_i \vk_{i,r'} \vq_{i,r'}^\T \vx_q \right) \vq_{i,r}^\T \vx_q \right]  \\
&= \vzero  .
\end{align*}
\end{proof}

\subsection{Gradient Flow Equations}
Based on the gradient flow training rule in \cref{eq:grad-flow}, the gradient flow dynamics of linear attention with separate rank-$R$ key and query is
\begin{subequations}  \label{eq:attnS-lowrank-gd}
\begin{align}
\tau \dot v_i &= \sum_{r=1}^R \vk_{i,r}^\T \E \left( \vbeta (y_q-\hat y_q) \vx_q^\T \right) \vq_{i,r}
= \sum_{r=1}^R \vk_{i,r}^\T \left( \mLambda^2 - \E\left({\hat\mLambda}^2\right) \sum_{i=1}^H \sum_{r'=1}^R v_i \vk_{i,r'} \vq_{i,r'}^\T \mLambda \right) \vq_{i,r}  ,\\
\tau \dot \vk_{i,r} &= v_i \E \left( \vbeta (y_q-\hat y_q) \vx_q^\T \right) \vq_{i,r}
= v_i \left( \mLambda^2 - \E\left({\hat\mLambda}^2\right) \sum_{i=1}^H \sum_{r'=1}^R v_i \vk_{i,r'} \vq_{i,r'}^\T \mLambda \right) \vq_{i,r}  ,\\
\tau \dot \vq_{i,r} &= v_i \vk_{i,r}^\T \E \left( \vbeta (y_q-\hat y_q) \vx_q \right)
= v_i \left( \mLambda^2 - \mLambda \sum_{i=1}^H \sum_{r'=1}^R v_i \vq_{i,r'} \vk_{i,r'}^\T \E\left({\hat\mLambda}^2\right) \right) \vk_{i,r}  .
\end{align}
\end{subequations}
where $i=1,\cdots,H,\, r=1,\cdots,R$ , and the data statistics $\E\left({\hat\mLambda}^2\right)$ is calculated in \cref{eq:ELambdaLambda}.

\subsection{Fixed Points  \label{supp:attnS-lowrank-landscape}}
We use $\gM(\gS_m)$ to denote a set of fixed points that correspond to learning $m$ ($m=0,1,\cdots,D$) out of the $D$ eigenvectors,
\begin{align}  \label{eq:attnS-lowrank-def-M}
\gM(\gS_m)= \left\{ v_{1:H}, \mW^K_{1:H}, \mW^Q_{1:H} \bigg| \text{conditions (C1)-(C3) are met} \right\}  ,
\end{align}
where the set $\gS_m$ specifies the indices of the learned eigenvectors,
\begin{align}
\gS_m \subseteq \{1,2,\cdots,D\} ,\, | \gS_m | = m .
\end{align}
The three conditions for \cref{eq:attnS-lowrank-def-M} are:
\begin{enumerate}[label=(C\arabic*)]
    \item The heads sum up to fit the eigenvectors with indices $\gS_m$
    \begin{align}
        \sum_{i=1}^H \sum_{r=1}^R v_i \vk_{i,r} \vq_{i,r}^\T = \sum_{d\in \gS_m} \lambda_d^{-1} \left( 1 + \frac{1+\tr(\mLambda)/\lambda_d}N \right)^{-1} \ve_d \ve_d^\T.
    \end{align}
    \item For heads with a nonzero value weight, $v_i\neq0$, $\vk_{i,r}, \vq_{i,r} \, (r=1,\cdots,R)$ all lie in the span of $\{ \ve_d \}_{d\in \gS_m}$.
    \item For heads with a zero value weight, $v_i=0$, 
    \begin{align}
        \sum_{r=1}^R \sum_{d \notin \gS_m} \lambda_d^2 \vk_{i,r}^\T \ve_d \ve_d^\T \vq_{i,r} = 0.
    \end{align}
\end{enumerate}
With the same reasoning as \cref{supp:attnS-landscape}, one can show the weights satisfying these three conditions have zero gradients and thus are fixed points. 
Though conditions (C1,C3) do not explicitly specify the weights, they are feasible conditions. One possible weight configuration that satisfies all three conditions is to let $\vk_{i,r},\vq_{i,r} \,(r\neq 1)$ be zero and let $v_i,\vk_{i,1},\vq_{i,1}$ be the same as the fixed point for linear attention with rank-one key query, where the low-rank case falls back into the rank-one case.
Therefore, the fixed points described in \cref{eq:attnS-lowrank-def-M} are valid and feasible.
Linear attention with separate rank-$R$ key and query has the same $2^D$ fixed points in the function space as its rank-one counterpart.

\subsection{Saddle-to-Saddle Dynamics  \label{supp:attnS-lowrank-explain-s2s}}

\begin{figure}[b]
  \centering
  \subfloat[$R=2$  ]
    {\centering
    \includegraphics[width=0.288\linewidth]{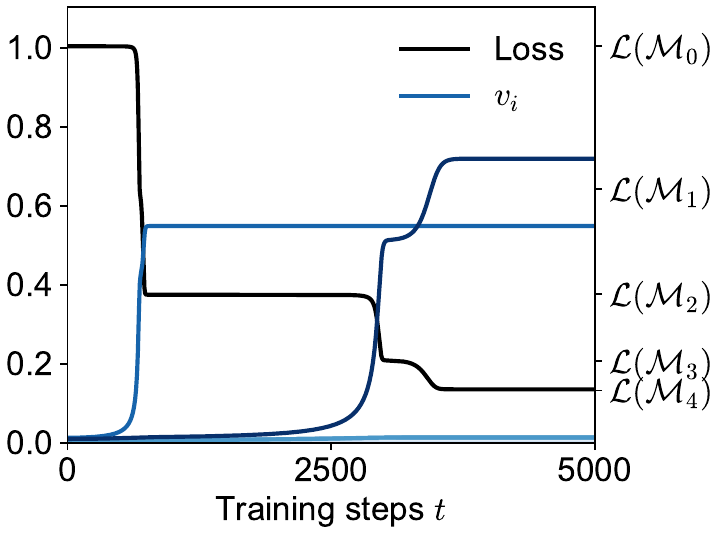}}
  \hspace{2ex}
  \subfloat[$R=3$  ]
    {\centering
    \includegraphics[width=0.288\linewidth]{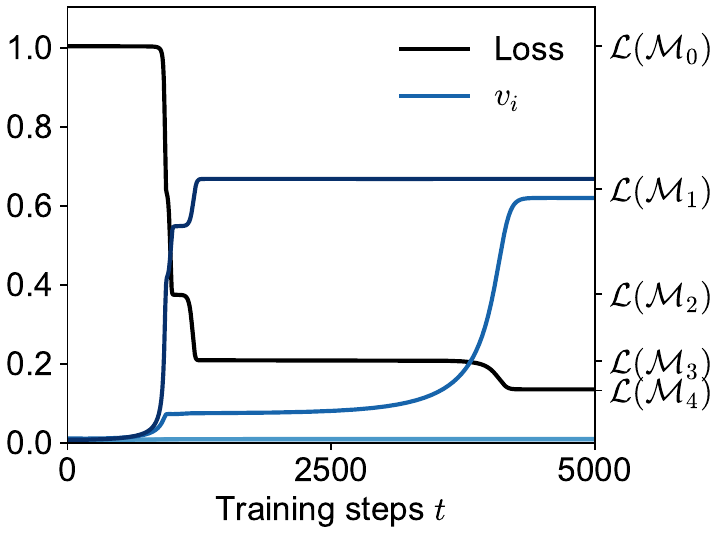}}
  \hspace{2ex}
  \subfloat[$R=4$  ]
    {\centering
    \includegraphics[width=0.288\linewidth]{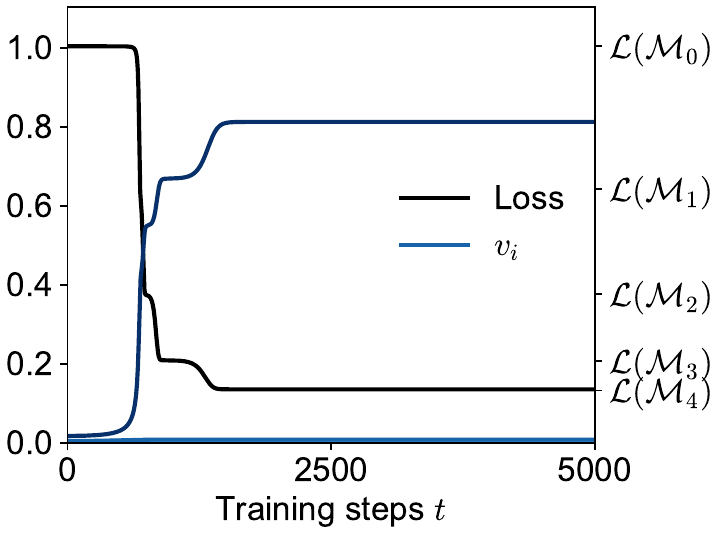}}
  \caption{Loss and value weights trajectories. The setting is the same as \cref{fig:attnS-value} except different ranks $R=2,3,4$. In the rank-one case in \cref{fig:attnS-value}, value weights in four heads grow, each corresponding to an abrupt loss drop from $\Ls(\gM_m)$ to $\Ls(\gM_{m+1}) \, (m=0,1,2,3)$. In the rank-$R$ case, a new value weight grows big from small initialization when the loss decreases from $\Ls(\gM_m)$ to $\Ls(\gM_{m+1})$ for $m$ that divides $R$. Here $D=4,N=31,H=5$, and $\mLambda$ has eigenvalues $0.4,0.3,0.2,0.1$.}
  \label{fig:attnS-lowrank-value}
\end{figure}

\begin{figure}
  \centering
    \includegraphics[width=\linewidth]{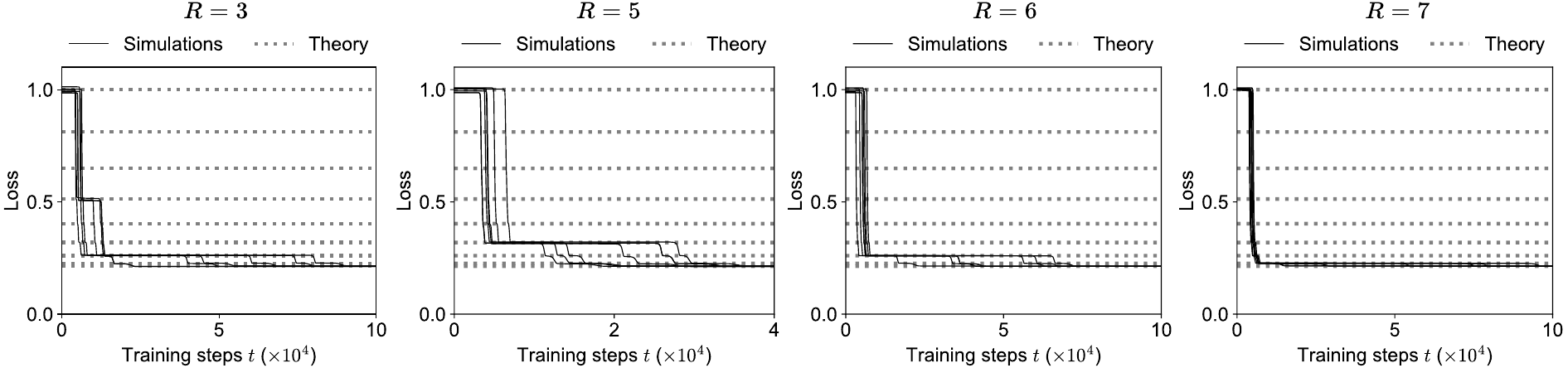}
  \caption{Same as \cref{fig:attnS-lowrank} but with ranks $R=3,5,6,7$. Here $D=8,N=31,H=9$, $\mLambda$ has trace $1$ and eigenvalues $\lambda_d\propto d^{-1}$.}
  \label{fig:attnS-lowrank-supp}
\end{figure}

For linear attention with rank-$R$ key and query, the gradient updates of the key and query weights in \cref{eq:attnS-lowrank-gd}, $\dot \vk_{i,r}, \dot \vq_{i,r}$, include the factor $v_i$, which is the shared across ranks $r=1,\cdots,R$ but unique to each head.
In linear attention with rank-one key and query initialized with small weights, the weights in a head, $v_i,\vk_i,\vq_i$, escape from the unstable zero fixed point to drive the first abrupt drop of loss.
Similarly, in the rank-$R$ model, the value weight $v_i$ and a pair of key and query weights $\vk_{i,r},\vq_{i,r}$ in a head escape from the zero fixed point to drive the first abrupt drop of loss.

However, the subsequent dynamics differ between the the rank-one and rank-$R$ models.
In the rank-one model, the loss will undergo a conspicuous plateau until weights in a new head, $v_{i'},\vk_{i'},\vq_{i'} \, (i'\neq i)$, escape from the zero fixed point to grow.
By contrast, in the rank-$R$ model ($R>1$), the loss will plateau briefly or not plateau because a new pair of key and query weights in the same $i$-th head, $\vk_{i,r'},\vq_{i,r'} \, (r'\neq r)$, can quickly grow to drive the loss drop. A new pair of key and query weights in the $i$-th head grows faster than the key and query weights in a new head, because the value weight in the $i$-th head, $v_i$, has already grown during the first abrupt loss drop. Since the gradient updates of all key and query weights in the $i$-th head include the factor $v_i$, a larger value weight leads to larger gradient updates for the associated key and query weights. We plot the value weights with $D=4$ and ranks $R=1,2,3,4$ in \cref{fig:attnS-value,fig:attnS-lowrank-value} to show: the loss drop after a conspicuous plateau corresponds to a new value weight escaping from zero, while the loss drop after a brief plateau does not.

We plot the loss trajectories with $D=8$ and different ranks in \cref{fig:attnS-lowrank-supp} to complement \cref{fig:attnS-lowrank} in the main text.

\subsection{Dynamics with Repeated Eigenvalues}
\begin{figure}[h]
  \centering
  \includegraphics[width=\linewidth]{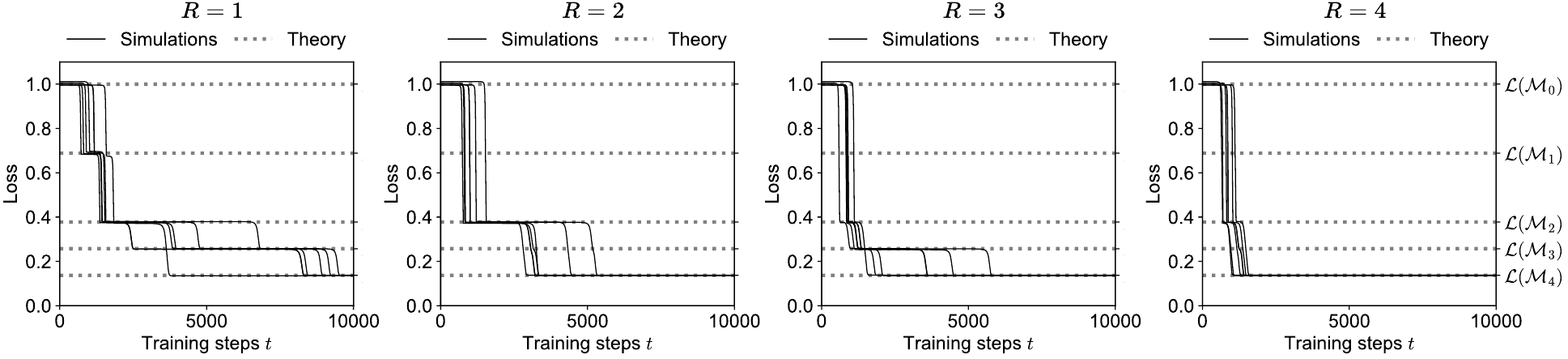}
  \caption{Loss trajectories of multi-head linear attention with separate key and query. The setup is the same as in \cref{fig:attnS-loss} except that $\mLambda$ has eigenvalues $0.35,0.35,0.15,0.15$. The four panels differ only in the rank of the key and query weights. Although some eigenvalues are equal, the loss trajectory of $\attn_{\text S}$ with $R=1$ can still exhibit plateaus when learning them, due to the different random initial weights in each head. The plateaus may also be skipped for certain random seeds.}
  \label{fig:attnS-loss-eqeig}
\end{figure}

We have demonstrated that linear attention with separate key and query exhibits loss plateaus during training when the eigenvalues of the input token covariance matrix, $\mLambda$, are distinct. When $\mLambda$ has repeated eigenvalues, linear attention with separate key and query can also exhibit loss plateaus due to the different random initial weights in each head. 
In the case with distinct eigenvalues (\cref{fig:attnS-loss}), the plateau duration is determined by both the size of the eigenvalues and the random initialization. In the case with repeated eigenvalues (\cref{fig:attnS-loss-eqeig}, leftmost panel), the plateau duration is determined solely by the random initialization.

\subsection{Conservation Law}
The gradient flow dynamics of linear attention with separate key and query in \cref{eq:attnS-lowrank-gd} implies a conservation law.
The value, key, and query weights in a head obey
\begin{align}  \label{eq:attnS-lowrank-conservation}
\frac{\diff}{\diff t} \left( \vk_{i,r}^\T \vk_{i,r} - \vq_{i,r}^\T \vq_{i,r} \right) = \vzero ,\quad
\frac{\diff}{\diff t} \left( \sum_{r=1}^R \vk_{i,r}^\T \vk_{i,r} - v_i^2 \right) = 0 .
\end{align}

We here prove that \cref{eq:attnS-lowrank-conservation} holds regardless of the choice of the loss function.
\begin{proof}
We can use the generic gradient flow equation in \cref{eq:grad-flow} to calculate the relevant gradients 
{\allowdisplaybreaks
\begin{subequations}
\begin{align}
\frac{\diff \vk_{i,r}^\T \vk_{i,r}}{\diff t} &= 2 \vk_{i,r}^\T \frac{\diff \vk_{i,r}}{\diff t} 
= 2 \E \left( - \vk_i^\T \frac{\diff \Ls}{\diff \hat y_q} \frac{\diff \hat y_q}{\diff \vk_{i,r}} \right)
= 2 \E \left( - \frac{\diff \Ls}{\diff \hat y_q} v_i \vk_{i,r}^\T \vbeta \vq_{i,r}^\T \vx_q  \right)
\label{eq:attnS-lowrank-kk}  \\
\frac{\diff \vq_{i,r}^\T \vq_{i,r}}{\diff t} &= 2\vq_{i,r}^\T \frac{\diff \vq_{i,r}}{\diff t} 
= 2 \E \left( - \vq_i^\T \frac{\diff \Ls}{\diff \hat y_q} \frac{\diff \hat y_q}{\diff \vq_{i,r}} \right) 
= 2 \E \left( - \frac{\diff \Ls}{\diff \hat y_q} v_i \vq_{i,r}^\T \vx_q \vk_{i,r}^\T \vbeta \right) 
\label{eq:attnS-lowrank-qq}  \\
\frac{\diff v_i^2}{\diff t} &= 2 v_i \frac{\diff v_i}{\diff t}
= 2 \E \left( - v_i \frac{\diff \Ls}{\diff \hat y_q} \frac{\diff \hat y_q}{\diff v_i} \right) 
= 2 \sum_{r=1}^R \E \left( - \frac{\diff \Ls}{\diff \hat y_q} v_i \vbeta^\T \vk_{i,r} \vq_{i,r}^\T \vx_q \right) 
\label{eq:attnS-lowrank-vv}
\end{align}
\end{subequations}
}
Comparing \cref{eq:attnS-lowrank-kk,eq:attnS-lowrank-qq}, we see that the following holds regardless of the specific choice of the loss function $\Ls$
\begin{align*}
\frac{\diff \vk_{i,r}^\T \vk_{i,r}}{\diff t} = \frac{\diff \vq_{i,r}^\T \vq_{i,r}}{\diff t}  .
\end{align*}
Similarly, comparing \cref{eq:attnS-lowrank-kk,eq:attnS-lowrank-qq} with \cref{eq:attnS-lowrank-vv}, we obtain
\begin{align*}
\sum_{r=1}^R \frac{\diff \vk_{i,r}^\T \vk_{i,r}}{\diff t} = \frac{\diff v_i^2}{\diff t}  .
\end{align*}
\end{proof}

\section{Training Dynamics of In-Context and In-Weight Learning  \label{supp:icl-iwl}}
\begin{figure}
  \centering
    \subfloat[$0\%$ fixed task (in-context)  \label{fig:icl-iwl-p0}]
    {\centering
    \includegraphics[width=0.3\linewidth]{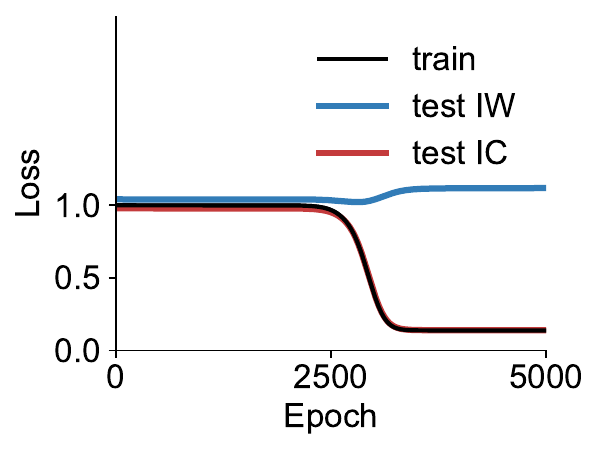}}
    \subfloat[$20\%$ fixed tasks]
    {\centering
    \includegraphics[width=0.3\linewidth]{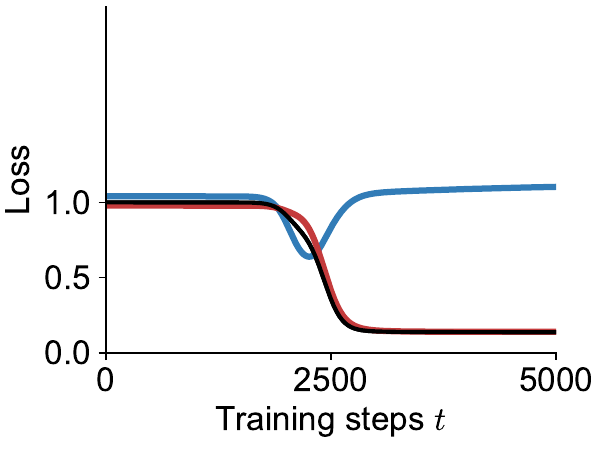}}
    \subfloat[$40\%$ fixed tasks]
    {\centering
    \includegraphics[width=0.3\linewidth]{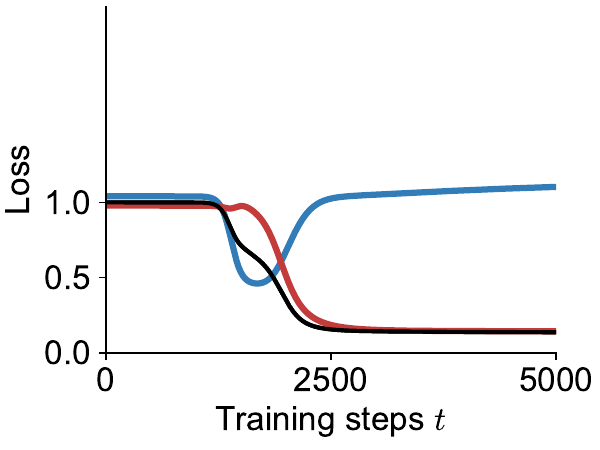}}
  \\
    \subfloat[$60\%$ fixed tasks]
    {\centering
    \includegraphics[width=0.3\linewidth]{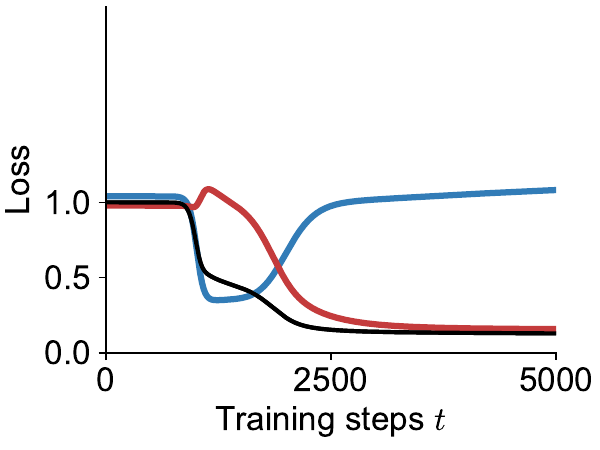}}
    \subfloat[$80\%$ fixed tasks]
    {\centering
    \includegraphics[width=0.3\linewidth]{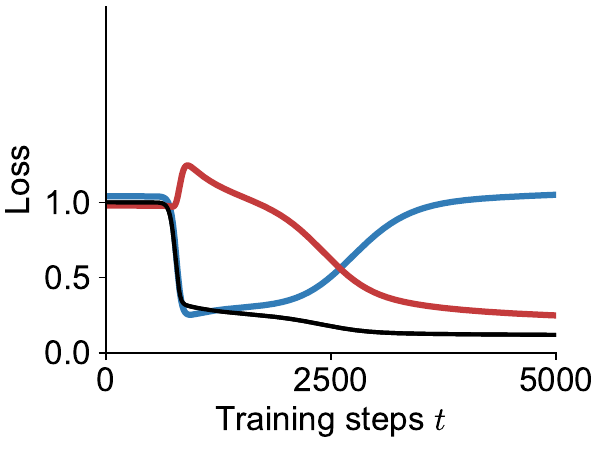}}
    \subfloat[$100\%$ fixed tasks (in-weight)]
    {\centering
    \includegraphics[width=0.3\linewidth]{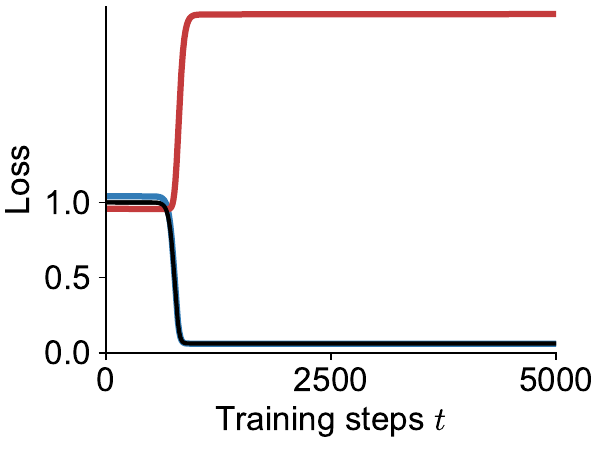}}
  \caption{Dynamics of in-context and in-weight learning in linear attention with merged key and query. The training set is the same as the in-context linear regression task described in \cref{sec:def-data} except that a portion of the task vectors $\vw$ are fixed. The portion of fixed task vectors indicates how much training samples can be fitted with the in-weight learning solution, that is memorizing the fixed task vector. The in-context learning test loss is evaluated on test sequences whose task vectors are all sampled from $\mathcal{N}(\vzero,\mI)$. The in-weight learning test loss is evaluated on test sequences whose task vector is the same fixed task vector from the training set. Here $D=4,N=31,H=8,\mLambda=\mI/D$.}
  \label{fig:icl-iwl-sweep}
\end{figure}

In this work, we focused on the training dynamics of ICL abilities. Other than ICL, attention models can also learn in weight, that is solving the task by memorizing the map between the query input and the target output without using the information in context. The arbitration between in-context and in-weight learning may depend on the properties of the training data \citep{chan22icliwl}.
To focus on the dynamics of ICL, we considered a purely ICL task, which is in-context linear regression with the task vector sampled from a zero-mean standard normal distribution, $\vw \sim \mathcal N(\vzero,\mI)$. Since memorizing any particular task vector does not effectively decrease the loss, linear attention develops only ICL ability during training, as shown in \cref{fig:icl-iwl-p0}.

If the task vector $\vw$ follows a different distribution, the training dynamics involves the development of both in-context and in-weight learning abilities. In \cref{fig:icl-iwl-sweep}, we let the task vector for some of the training sequences be fixed and sample the rest from a standard normal distribution to elicit in-weight learning ability. We plot the training loss, in-context learning test loss, and in-weight learning test loss for varying portions of fixed task vectors in \cref{fig:icl-iwl-sweep}. The larger the portion of fixed task vectors, the lower the loss the model can achieve by memorizing the fixed task vector in weight. We indeed observe the training loss and in-weight learning test loss are lower right after the first abrupt loss drop when the portion is larger.
% Unlike \cref{fig:icl-iwl-p0}, the dynamics of in-context and in-weight learning dynamics interact and possibly compete when there are training sequences with a fixed task vector.

% When fixing a portion of the task vectors and sampling the rest from $\mathcal N(\vzero,\mI)$, we find that the linear attention model first learns in weight and then learns in context, as shown in \cref{fig:icl-iwl-sweep}. We plot the training loss and the test loss trajectories with varying portions of fixed task vectors in \cref{fig:icl-iwl-sweep}. 

% For the in-context linear regression task we focused on in the main text, we sample the task vectors for all training sequences from a standard normal distribution, $\vw \sim \mathcal{N}(\vzero,\mI)$. In this case, the linear attention model only develops in-context learning ability as shown in \cref{fig:icl-iwl-p0}.

The technical consequence of fixing some of the task vectors is that \cref{eq:attnM-zerov,eq:attnM-zerou} break. In other words, we cannot assume the certain blocks of the value and the merged key-query matrices are zero as in \cref{supp:attnM-zerouv}. Without the zero block assumption, the linear attention model implements
\begin{align}  \label{eq:attnM-icliwl}
\hat y_q = \sum_{i=1}^H \left( \vv_i^\T \left( \hat \mLambda + \frac1N \vx_q \vx_q^\T \right) \mU_i + v_i \vbeta^\T \mU_i + \vv_i^\T \vbeta \vu_i^\T + v_i \frac1N \sum_{n=1}^N y_n^2 \vu_i^\T \right) \vx_q  .
\end{align}
\cref{eq:attnM-icliwl} include not only a linear map of the cubic feature $\vz=\VEC(\vbeta \vx_q^\T)$ but also linear maps of additional features, $\left( \hat \mLambda + \frac1N \vx_q \vx_q^\T \right) \otimes \vx_q, \frac1N \sum_{n=1}^N y_n^2 \vx_q$. Future work could analyze the gradient descent dynamics of the model described by \cref{eq:attnM-icliwl}, building on our results on the dynamics of in-context learning to explore its interactions with in-weight learning.

